\pdfoutput=1

\documentclass[11pt]{article}

\usepackage[final]{acl}

\usepackage{authblk}

\usepackage{times}
\usepackage{latexsym}

\usepackage[T1]{fontenc}
\usepackage[utf8]{inputenc}
\usepackage{microtype}

\usepackage{inconsolata}

\usepackage{fontawesome}

\usepackage{amssymb}
\usepackage{amsmath}
\usepackage{bm}

\usepackage{graphicx}

\graphicspath{{./figures}}

\usepackage{tabularray}
\UseTblrLibrary{booktabs}
\UseTblrLibrary{siunitx}

\usepackage[hang,flushmargin]{footmisc}

\usepackage[inline]{enumitem}
\usepackage{cleveref}

\newcommand{\Ni}{(1)~}
\newcommand{\Nii}{(2)~}
\newcommand{\Niii}{(3)~}
\newcommand{\Niv}{(4)~}


\title{When Attention Goes Blind:\\Numerical Failure in ALiBi Positional Encodings}

\author[1,2]{Christopher Schröder}
\author[3,2]{Lukas Gienapp}
\author[3]{Ferdinand Schlatt}
\author[4,5,2]{\authorcr Martin Potthast}
\author[1]{Gerhard Heyer}

\affil[1]{Institute for Applied Informatics at Leipzig University (InfAI)}
\affil[2]{ScaDS.AI Dresden/Leipzig}
\affil[3]{Seltz}
\affil[4]{University of Kassel}
\affil[5]{hessian.AI}

\begin{document}
\maketitle
\begin{abstract}
We identify a previously overlooked failure mode of ALiBi positional encoding: its linearly growing bias eventually underflows floating-point precision, which zeroes out a large fraction of attention weights and renders the affected attention heads partially blind. We analyze this failure mode, formalize the distance at which each head goes blind, and confirm its occurrence in pretrained ALiBi models. To assess its impact, we pretrain 148M-parameter decoder models with altered slopes and four mitigation strategies, individually and in combination. We find that preventing underflow by clamping leaves performance on downstream benchmarks and associative retrieval unchanged, indicating that underflow does not necessarily degrade ALiBi models. In contrast, log-scaled distances with explicit slopes substantially improve associative retrieval, especially beyond the training context. Based on these findings, we provide concrete recommendations on how to train models with ALiBi.%
\hspace{-0.2em}\footnote{Code: \href{https://github.com/coral-nlp/eacl2027-when-attention-goes-blind}{\raisebox{-0.11em}{\normalsize\faGithub}~coral-nlp/eacl2027-when-attention-goes-blind}.}
\end{abstract}

\section{Introduction}

\begin{figure}[!t]
\centering
\includegraphics[width=\linewidth]{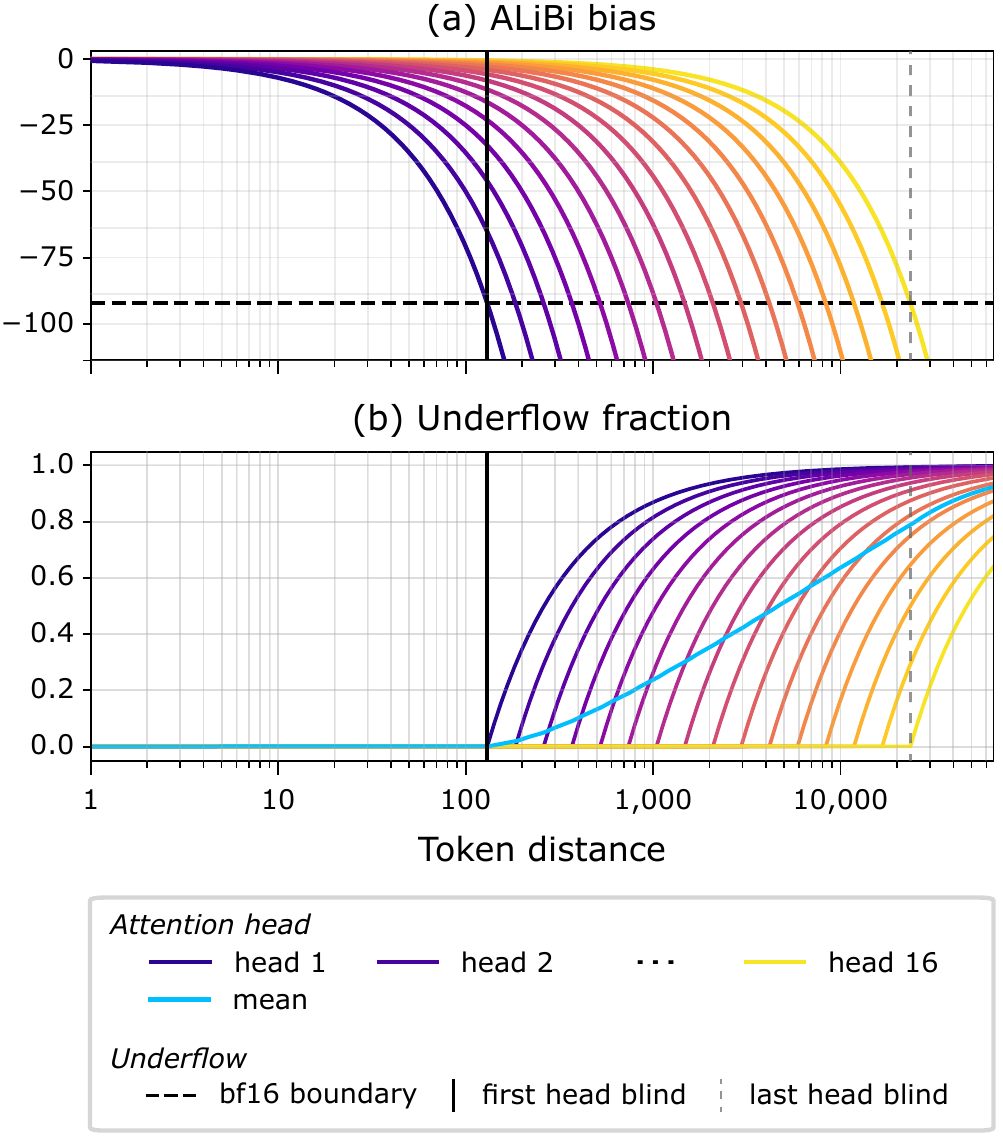}
\caption{ALiBi bias and fraction of underflowed attention weights for a 16-head layer in bf16 precision: (a)~ALiBi encodes position through an additive bias (y-axis) proportional to token distance and a head-specific slope (x-axis). Attention weights underflow when the bias crosses a threshold (dashed line). (b)~Each underflow zeroes a weight in the self-attention matrix, {\em blinding} heads from those token positions onward.
}
\label{fig:bias-and-underflow}
\end{figure}

Positional encoding through {\em attention with linear biases} (ALiBi; \citealt{press:2022}) is a computationally cheap, parameter-free, and memory-efficient approach to encode position information in the Transformer model. It is conditioned on the relative distance of token pairs applied as bias to attention weights. \citeauthor{press:2022} claim that, unlike other positional encodings, ALiBi enables straightforward extrapolation beyond the training context window.

We show, however, that in practice ALiBi does not behave as implied by its formal definition (Figure~\ref{fig:bias-and-underflow}). As a direct consequence of the unbounded linear scaling of the bias with increasing token distance, under finite floating-point arithmetic, attention weights will inevitably underflow beyond a distance determined by ALiBi's per-head {\em slope} hyperparameter. The underflow then zeroes affected attention scores and renders the respective attention head {\em blind} to interactions. Since this occurs at increasing token distances, it notably opposes ALiBi's promise of length extrapolation. Eventual underflow is inevitable, but the impact of token blindness on model performance remains unclear.

To the best of our knowledge, this problem has not been reported before.
We systematically investigate it with the following contributions: 
We
\begin{enumerate*}[label=(\arabic*)]
\item
analytically show the cause and immediate implications of attention weights underflowing due to ALiBi bias (\Cref{sec:alibi});
\item
assess existing pre-trained ALiBi models for evidence of this problem (\Cref{sec:task-failures});
\item
train decoder models with slopes set to induce or avoid attention blindness (\Cref{sec:training-failure-mode});
\item
and explore several training-level mitigation strategies (\Cref{sec:training-mitigations}).
\end{enumerate*}

We find that preventing underflow simply by clamping does not measurably improve ALiBi models. 
Multiple combinations of mitigation strategies, however, still are beneficial, especially those combining log-scaled distances with explicit slopes. The best combination, explicit slopes with log-scaled distances and soft capping (E+L+S), improves area under the accuracy curve for all three associative retrieval tasks, most notably beyond the training context, where it raises the area under curve from $0.05$ to $0.98$ for passkey retrieval, from $0.13$ to $0.58$ for single-key, and from $0.11$ to $0.38$ for multi-key needle in a haystack. 

\section{Related Work}

The Transformer architecture~\cite{vaswani:2017} processes all tokens in parallel, unaware of their sequential order. {\em Positional encodings} such as ALiBi inject word order into the model, e.g., by modulating token embeddings, rotating the query and key vectors, or biasing the attention logits.

We review positional encodings relevant to our work, the broader family of additive-bias positional encodings, their known limitations, and, closest to our work, failures due to numerical precision.

\subsection{Positional Encodings}

In conjunction with the attention mechanism, \citet{vaswani:2017} proposed fixed \emph{sinusoidal positional encodings}, which inject positional information when added onto the token embeddings.
They also investigated \emph{learned absolute positional encodings}, reporting them to be very close in effectiveness and dismissing them for sinusoidal embeddings which can extrapolate to some degree beyond the training context length.

\emph{ALiBi}~\cite{press:2022} instead adds a bias to the attention logits that grows with the distance between query and key, which costs little compute and lets attention operate beyond the training context length~\cite{zhao:2024b}.

\emph{Rotary Position Embedding} (RoPE; \citealt{su:2024})%
\footnote{\citet{su:2024} write \emph{embedding} where \citet{vaswani:2017} write \emph{encoding} for a comparable mechanism. We use the more general term encoding throughout.}
encodes the absolute position of a token as a position-dependent rotation of its query and key vectors.
The query--key dot product then depends only on the distance between the two positions, so RoPE also carries relative information.
RoPE is now the default positional encoding, and one explanation is that it outperformed ALiBi in later comparisons~\cite{li:2024,zhao:2024b}.
We return to the ALiBi side of that comparison.

\subsection{Positional Encoding Through Bias}

Few works follow up on ALiBi itself, but a line of research keeps the idea of encoding position as an additive bias.
\emph{KERPLE}~\cite{chi:2022} generalizes the linear bias of ALiBi to a family of kernels with learnable parameters, including log scaling and power scaling.
\emph{FIRE}~\cite{li:2024} learns the bias as a function of the relative position normalized by the sequence length, with the aim of extrapolating to longer contexts.
\emph{Sandwich}~\cite{chi:2023} derives a relative bias from the absolute sinusoidal encodings and applies it after the softmax.
To our knowledge, none of these biases is used outside research.
One explanation is that each adds computation ALiBi does not need, leaving ALiBi the cheapest of these additive-bias encodings and the one this paper studies.

\subsection{Limitations of Positional Encodings}

For ALiBi, previous work observed the supposedly strong extrapolation capabilities to underperform~\cite{chi:2023,kazemnejad:2023,chen:2024,zhu:2024b}.
Apart from that, \citet{schallon:2026} identified a systematic collapse of attention heads where a notable fraction of attention heads attends predominantly to the begin-of-sequence token. This is distinct from the numerical deficiency that we study.

Similar problems, however, also exist for RoPE:~\citet{barbero:2025} find that the Gemma model~\cite{gemmateam:2024} stores semantic information in the low rotation frequencies, whose signal weakens at longer contexts.
\citet{du:2026} show that RoPE loses the ability to discriminate both position and token as the context grows.

While positional encodings have become recognized as a crucial component of Transformers, central problems such as limited extrapolation capabilities persist.
Our work discovers and investigates a systematic deficiency in ALiBi, which could help to explain when and why extrapolation fails.

\subsection{Failure due to Numerical Precision}

Numerical issues affecting the correctness of established functions or even larger algorithms have recently gained increasing attention: \citet{blanchard:2021} investigate different softmax formulations in terms of floating-point accuracy and find that underflow is generally harmless with regard to softmax correctness.
\citet{barron:2026} recently identified two numerical bugs in established implementations of softmax and log-softmax which resulted in incorrect gradients.
This emphasizes that revisiting numerical correctness is crucial.

In contrast to these works, our problem manifests at a higher level within ALiBi-based attention. At the same time, it is narrower in scope, affecting only models that use ALiBi.
Closest to our work, \citet{wang:2025} trace a failure in RoPE to numerical precision: under bfloat16 precision, the relative position that RoPE encodes is compromised.
To the best of our knowledge, this is the only prior work that investigates numerical failures in positional encodings, and none exists for ALiBi.
\section{Numerical Failure in ALiBi} 
\label{sec:alibi}

ALiBi adds a per-head bias to the attention logits that grows linearly with the distance between query and key, which costs little compute and lets attention operate beyond the training context length~\cite{zhao:2024b}.
We first restate ALiBi in our notation, then show that its bias drives the softmax into underflow.

\subsection{Attention with Linear Biases}

All of the following describes a single attention head with a per-head scaling factor $m$, called {\em slope}. We drop the head index $h$ and token indices $i, j$ where the context makes them clear.

ALiBi additively combines attention logits with a bias matrix $B$:
\begin{equation}
\mathrm{softmax}\!\left(\frac{QK^\top}{\sqrt{d_k}} + B\right)V.
\label{eq:attn}
\end{equation}

\noindent 
We write the attention logits as
\begin{equation*}
A = \frac{QK^\top}{\sqrt{d_k}}.
\end{equation*}

\noindent The bias $B$ is the product of the negative slope and the token--token distance matrix $D$:
\begin{equation}
B = -m \cdot D,\ \mbox{where}
\label{eq:alibi-bias}
\end{equation}
\begin{equation}
D = (i - j)  \cdot \mathbf{1}[j < i]
\end{equation}

\noindent \citet{press:2022} fix the slopes to a geometric sequence shared across all layers. Most published ALiBi models keep that choice (Appendix Table~\ref{table:models}).

\subsection{Softmax Underflow}

The failure occurs when the exponentials inside the softmax of Equation~\ref{eq:attn} fall below the smallest positive value that the floating-point format can represent. Recall the softmax definition:
\begin{equation}
\mathrm{softmax}\!\left(\mathbf{x}\right)_i = \frac{e^{x_i}}{\sum_{l=1}^{d}e^{x_l}}.
\label{eq:softmax}
\end{equation}

\noindent 
In row $i$ of the attention matrix, the exponent of column $j$ is $A_{i,j} + B_{i,j}$. As the distance $D_{i,j}$ grows, the bias dominates the logits and pushes the exponent toward more negative values. Under numerical precision limits, a threshold $\tau_{u} < 0$ exists below which the exponential function underflows,\hspace{-0.2em}\footnote{Some implementations rewrite this into the equivalent formulation $\mathrm{softmax}\!\left(A + m_hj\right)$ which is nevertheless prone to underflows. See Appendix Section~\ref{app:softmax-invariance} for details.} i.e., $e^{\tau_{u}} = 0$. 
The logits $A$, the bias $B$, or their sum can push the exponent past this threshold. The logits vary in magnitude with the query and key vectors, which are functions of the input. The bias does not depend on the input and grows with token distance, and therefore is guaranteed to cross the threshold eventually. Figure~\ref{fig:bias-and-underflow}, shows how the bias grows at increasing token distances for default slopes, as well as the resulting underflow fraction under the simplifying assumption of $A = 0$. The thresholds\footnote{See Appendix Section~\ref{app:floating-point-thresholds} on floating-point thresholds.} of the two dominant formats are $\tau_{u,\mathrm{fp32}} \approx -103.28$ and $\tau_{u,\mathrm{bf16}} \approx -92.19$.

Heads are constructed with different slopes to model different attenuation to token distance~\cite{press:2022}, typically monotonically decreasing. Thus, the head with the steepest slope then underflows at the smallest distance and the head with the flattest slope at the largest, which we call \emph{partial} and \emph{full} attention blindness.

For each head $h$, the \emph{blindness distance} $\Delta_h(\varepsilon)$ is the smallest
distance at which an entry with attention logit $\varepsilon = A_{i,j}$ underflows:
\begin{align*}
    \Delta_h(\varepsilon) &= \min\{d \in \mathbb{N} \mid \varepsilon - m_h d \le \tau_u\}\\
    &= \left\lceil \frac{\varepsilon - \tau_u}{m_h} \right\rceil \quad (\text{for } \varepsilon > \tau_u),\\ 
    \delta_h :&= \Delta_h(0).
\end{align*}

Here, $\tfrac{\varepsilon}{m_h}$ is inversely proportional to the slope, so the flat-slope heads move most under real logits. We write $\delta_h$ for the simplified case
$\varepsilon = 0$, under which $\delta_1$ marks the onset of partial blindness and
$\delta_H$ the onset of full blindness.

\subsection{Consequences of Underflow}

Underflow has two consequences, both visible in Equation~\ref{eq:softmax}:
\Ni~The primary effect is \emph{positional blindness}: as a result of the underflow, the numerator becomes zero, so every token beyond $\delta_h$ receives an attention weight of zero and cannot contribute to the values $V$. 
\Nii~The zeroed terms also leave the denominator, which redistributes weight over the remaining tokens in the same row.
We call this second effect \emph{weight redistribution}.
Heads with steep slopes go blind early and act as a sliding window. \citet{chi:2023} already observe the sliding window behavior, but do not report the underflow behind it.

Figure~\ref{fig:bias-and-underflow}b shows how the fraction of blinded entries in the attention matrix grows with token distance.
For a query at position 2{,}048, 35.6\% of its attention entries (averaged over heads) have crossed the underflow threshold.
Non-zero logits shift these points to the left or to the right, depending on the data, but any exponent that scales linearly with token distance underflows once the context is long enough. \emph{Whether and to what extent this affects model training and inference remains unclear}.

\subsection{Preventing Underflow}
\label{sec:fixes}

Is this underflow unavoidable? 
We investigate four mutually non-exclusive strategies which are intended to prevent or alleviate the failure.

\paragraph{Clamping} This strategy clamps the bias above the smallest value floating point can represent:
\begin{equation*}
B_{i,j}^{*} = \max(B_{i,j}, c_{clamp}).
\end{equation*}

\noindent Where $B_{i,j}$ is greater than or equal to $c_{clamp}$, the ALiBi bias behaves as before. Below it, the model treats all tokens as if they had the same distance, which loses positional information but prevents bias-induced underflow and weight redistribution.

\paragraph{Explicit Slopes} A linearly scaling bias always underflows at some distance, so the second strategy chooses the slopes such that the blindness distances spread across the context length of the model, e.g., adapted to a maximum context length of 2{,}048:
\begin{equation*}
\delta_{1} \overset{!}{=}32,\quad
\delta_{2} \overset{!}{=}64,\quad ...,\quad
\delta_{H} \overset{!}{=}2{,}048\quad
\end{equation*}

\noindent The schedule is flexible but depends on training context length among other factors. We further investigate this in Section~\ref{sec:training-failure-mode}.

\paragraph{Log-scaled Distances} The third strategy compresses the distance matrix with a logarithm, which delays the point at which the bias passes $\tau_u$:
\begin{equation*}
D_{i,j}^{*} = \log(D_{i,j} + 1).
\end{equation*}

\noindent For the default ALiBi slopes and assuming $A = 0$, log scaling moves $\delta_{1}$ from $124$ to $\num{4.37e53}$, which puts underflow out of reach for a large majority of contemporary models' context sizes.
This closely corresponds to the logarithmic variant of KERPLE~\cite{chi:2022} but without its two learnable parameters.

\paragraph{Soft Capping} The fourth strategy constrains the attention logits $A_{i,j}$ to a fixed interval  $[-z, z]$ through 
soft capping~\cite{bello:2016,riviere:2024}:
\begin{equation*}
A_{i,j}^{*} = z \cdot \tanh(\frac{A_{i,j}}{z}).
\end{equation*}

\noindent We cap the logits rather than the bias, which bounds how far the logits push the exponent and keeps the linear scaling of the bias intact.

\section{Experiments}

\begin{table*}[!bth]%
\centering%
\small
\SetTblrInner{rowsep=1pt,colsep=8pt}
\begin{tblr}{
  colspec={@{}p{2cm}@{}rl
  rrr
  p{1.2cm}
  @{}
  p{1.05cm}
  @{}
  p{0.5cm}
  @{}
  p{0.6cm}
  rr@{}},
} 
\toprule
{\bfseries Model} & \textbf{Size} & \textbf{Language(s)}  & \textbf{Window} & \textbf{Layers} & \textbf{Heads} & \SetCell[c=4]{c} \textbf{Slopes} & & &  & \bm{$\delta_1$} & \bm{$\delta_H$}\\
\midrule
BLOOM & 560M & 46 lang.$^1$ & 2,048 & 24 & 16 & $2^{-0.50}$, & $ 2^{-1}$, & $ ...$, & $2^{-8}$ & 124 & 22,358\\
Falcon-RW & 7B & eng & 2,048 & 36 & 64 & $2^{-0.125}$, & $2^{-0.25}$, & $ ...$, & $ 2^{-8}$\hfill & 95 & 22,358\\
MPT & 7B & eng & 2,048 & 32 & 32 & $2^{-0.25}$, & $2^{-0.50}$, & $...$, & $2^{-8}$ & 104 & 22,358\\
\bottomrule
\end{tblr}
\caption{Selected pre-trained decoder models that were trained with ALiBi. These models are subject of the first experiment and differ in number of parameters, training context window, number of layers, number of attention heads, and slopes. The slopes mostly follow the standard geometric series of \citet{press:2022}.}
\label{table:experiment-pretrained-models}
\end{table*}

We systematically investigate this failure mode in three stages: we confirm its existence in pre-trained models (\Cref{sec:task-failures}), attempt to isolate it through modulating slopes (\Cref{sec:training-failure-mode}), and evaluate the proposed strategies for mitigation (\Cref{sec:training-mitigations}).

\begin{figure*}[!t]
\centering
\includegraphics[width=\textwidth]{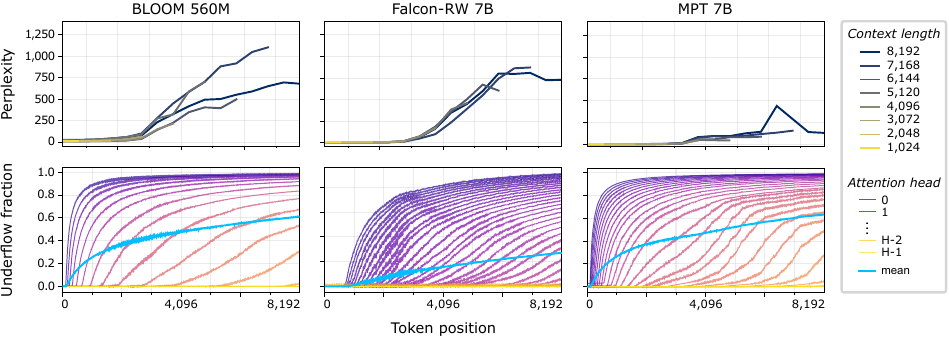}
\caption{Perplexity probes and empirical underflow for three pretrained ALiBi models with a training context of 2{,}048 tokens, evaluated on the length-stratified FineWeb-Edu test set.
}
\label{fig:perplexity-probe}
\end{figure*}

\subsection{Inspecting the Failure Mode}
\label{sec:task-failures}

To provide evidence for the analytically identified attention blindness, we investigate pre-trained models, since training could either resolve or exacerbate the failure mode, confounding the analysis.

\subsubsection{Setup}

The first experiment performs perplexity evaluations for causal language modeling, where we also  measure the fraction of underflowed tokens.

\paragraph{Models} We select three representative pre-trained decoder models across different model families and sizes: BLOOM~\cite{lescao:2022}, Falcon-RW~\cite{penedo:2023}, and MPT~\cite{mosaic:2023}.
Key characteristics, slopes, and expected blindness distances are listed in Table~\ref{table:experiment-pretrained-models}.

\subsubsection{Evaluation}

We conduct a perplexity test, during which we observe the test perplexity at increasing context lengths, and prompt for associative retrieval.

\paragraph{Data}

We sample a length-stratified test set from FineWeb-Edu~\cite{penedo:2024} for both the perplexity test and needle in a haystack probe. It consists of $32{,}000$ English documents evenly distributed across 32 bins of size $256$ up to a maximum context length of $8{,}192$~tokens.

\paragraph{Perplexity and Associative Retrieval}

While we can measure underflow, we cannot directly measure the impact of underflowed tokens.
Instead, we evaluate on three associative token retrieval tasks: passkey~\cite{mohtashami:2023}, single-key needle in a haystack~\cite{kamradt:2023}, and multi-key needle in a haystack~\cite{hsieh:2024}. All three are commonly used benchmarks in prior work~\cite{han:2024,hsieh:2024,peng:2024,an:2025,gelberg:2026}.

\emph{\Ni~Perplexity Probe and Underflow Fraction:}
\label{sec:attention-patterns}
To empirically check for the effect of the positional bias on causal language modeling, we compute perplexity on the aforementioned perplexity test set.
We also measure attention underflow to check for potential correlations with the perplexity curve.

\emph{\Nii~Passkey Retrieval:} In passkey retrieval~(PK; \citealp{mohtashami:2023}) a prompt begins by explicitly stating a passkey followed by an increasing amount of filler text.
We use word passphrases instead of numbers to adapt this task to smaller models, thereby decoupling numerical understanding from retrieval.

\emph{\Niii~Single-Key Needle in a Haystack} (NIAH$_s$; \citealp{kamradt:2023}): The needle in a haystack probe is a variation of passkey, where the password is put into an existing document.
In addition to the growing distance, the additional challenge here is to attend to the correct token among many distractors.

\emph{\Niv~Multi-Key Needle in a Haystack} (NIAH$_m$; \citealp{hsieh:2024}): In addition to the single-key variant by \citet{kamradt:2023}, we evaluate a more challenging multi-key variant from the RULER benchmark~\cite{hsieh:2024}, where the model has to return the value for one of multiple keys.

All retrieval probes are evaluated at increasing token distances between the target (passkey or needle) to prompt position.
This gives an accuracy curve over token distance which we evaluate from $128$ to $4{,}096$ (twice the training context length).
We report the area under this curve, separately within and beyond the training context.

\subsubsection{Results}

The results of the perplexity probe are shown in Figure~\ref{fig:perplexity-probe}.
First, for all three models, perplexity rises very slowly until a certain point is reached, which is always beyond the $2{,}048$-token training context.
Second, the empirically measured underflow fractions (averaged per head over all layers) look remarkably similar to the analytically constructed curve in Figure~\ref{fig:bias-and-underflow}b.

The results of the token retrieval probes are presented in Table~\ref{table:exp1-results}.
We see that even in-context retrieval is not perfect for the smaller BLOOM model.
NIAH scores are considerably lower than passkey, especially regarding out-of-context performance.

\begin{table}
\small
\centering
\SetTblrInner{rowsep=1pt,colsep=3pt}
\begin{tblr}{
  colspec   = {l@{\hspace{14pt}}r
    @{\hspace{10pt}}r
    @{\hspace{12pt}}r
    @{\hspace{10pt}}r
    @{\hspace{12pt}}r
    @{\hspace{10pt}}r},
  rowsep    = 2pt,
  colsep    = 8pt,
  leftsep = 0pt,
  rightsep = 0pt,
  row{1}    = {font=\bfseries},
}
\toprule
& \SetCell[c=2]{c} PK & & \SetCell[c=2]{c} NIAH$_s$ & & \SetCell[c=2]{c} NIAH$_m$ & \\
\cmidrule[lr]{2-3}\cmidrule[lr]{4-5}\cmidrule[lr]{6-7}
\bfseries Name & in & out & in & out & in & out \\

\midrule
BLOOM & 0.77 & 0.21 & 0.44 & 0.05 & 0.31 & 0.03\\
Falcon-RW & \textbf{1.00} & \textbf{0.51} & \textbf{0.62} & \textbf{0.10} & \textbf{0.58} & \textbf{0.07}\\
MPT & \textbf{1.00} & 0.50 & 0.51 & 0.09 & 0.50 & 0.06\\
\bottomrule
\end{tblr}
\caption{
\label{table:exp1-results}
Area under curve scores for passkey (PK), single-key (NIAH$_s$), and multi-key (NIAH$_m$) retrieval, within (in) and beyond the training context (out).
}
\end{table}

\subsubsection{Interim Discussion} 

The perplexity probes in Figure~\ref{fig:perplexity-probe} show notable spikes starting around $1.5$ times the training context size (of $2{,}048$).
The underflow fraction plots strongly resemble the analytically expected shape depicted in Figure~\ref{fig:bias-and-underflow}b.
There is no apparent correlation between perplexity and underflow fraction. 
Most importantly, \emph{we cannot infer whether these spikes appear due to exceeding the context, attention blindness, or both.}
Nevertheless, the measured underflow fractions empirically confirm the existence of the investigated deficiency.

The retrieval probes show notable differences across models. These differences confirm the retrieval probes as a sensitive criterion for the training experiments that follow.\hspace{-0.2em}\footnote{Distance-dependent curves in Appendix Section~\ref{app:retrieval-curves}.} 

\begin{table*}[!tbh]
\small
\centering
\begin{tblr}{
  colspec   = {@{}l@{\hspace{12pt}}rr@{\hspace{12pt}}rrrr@{\hspace{12pt}}rrr},
  rowsep    = 2pt,
  colsep    = 3.2pt,
  row{1,2}  = {font=\bfseries},
  column{1} = {leftsep=0pt},
  column{10} = {rightsep=0pt},
}
\toprule
 & \SetCell[c=6]{c} Associative Token Retrieval &  &  &  &  &  & \SetCell[c=3]{c} Downstream &  &  \\\cmidrule[lr]{2-7}\cmidrule[lr]{8-10}
Name & \SetCell[c=1]{c} PK$^{in}$ & \SetCell[c=1]{c} PK$^{out}$ & \SetCell[c=1]{c} NIAH$_s^{in}$ & \SetCell[c=1]{c} NIAH$_s^{out}$ & \SetCell[c=1]{c} NIAH$_m^{in}$ & \SetCell[c=1]{c} NIAH$_m^{out}$ & \SetCell[c=1]{c} CS & \SetCell[c=1]{c} QA & \SetCell[c=1]{c} Language \\
\midrule
\SetCell[c=10]{c} (a)~Impact of Slopes on the Failure Mode\\
\midrule
Steep & \underline{0.85$_{\pm 0.12}$} & \textbf{\underline{0.09$_{\pm 0.08}$}} & 0.45$_{\pm 0.06}$ & 0.01$_{\pm 0.01}$ & 0.25$_{\pm 0.04}$ & 0.00$_{\pm 0.00}$ & 47.1$_{\pm 0.7}$ & 46.1$_{\pm 0.5}$ & 52.4$_{\pm 0.4}$\\
Safe & \textbf{\underline{0.98$_{\pm 0.01}$}} & \underline{0.08$_{\pm 0.08}$} & 0.46$_{\pm 0.04}$ & 0.01$_{\pm 0.01}$ & 0.23$_{\pm 0.03}$ & 0.01$_{\pm 0.00}$ & 47.9$_{\pm 0.5}$ & \underline{47.0$_{\pm 0.6}$} & 53.2$_{\pm 0.2}$\\
Wide & \underline{0.91$_{\pm 0.13}$} & 0.04$_{\pm 0.03}$ & 0.47$_{\pm 0.03}$ & 0.01$_{\pm 0.01}$ & 0.24$_{\pm 0.06}$ & 0.01$_{\pm 0.00}$ & 48.6$_{\pm 1.6}$ & \textbf{\underline{\underline{47.9$_{\pm 0.3}$}}} & 53.8$_{\pm 0.2}$\\
ALiBi & 0.80$_{\pm 0.24}$ & 0.05$_{\pm 0.04}$ & \textbf{0.65$_{\pm 0.05}$} & \textbf{0.13$_{\pm 0.04}$} & \textbf{0.36$_{\pm 0.05}$} & \textbf{0.11$_{\pm 0.04}$} & \textbf{49.1$_{\pm 1.5}$} & 46.9$_{\pm 0.9}$ & 53.9$_{\pm 0.5}$\\
RoPE & 0.66$_{\pm 0.05}$ & 0.00$_{\pm 0.00}$ & 0.29$_{\pm 0.03}$ & 0.00$_{\pm 0.00}$ & 0.15$_{\pm 0.02}$ & 0.00$_{\pm 0.00}$ & 48.1$_{\pm 0.5}$ & \underline{47.1$_{\pm 0.7}$} & \textbf{\underline{\underline{54.4$_{\pm 0.5}$}}}\\
\midrule
\SetCell[c=10]{c} (b)~Mitigation Strategies\\
\midrule
Explicit (E) & \underline{0.95$_{\pm 0.05}$} & 0.03$_{\pm 0.04}$ & 0.50$_{\pm 0.03}$ & 0.03$_{\pm 0.02}$ & 0.32$_{\pm 0.02}$ & 0.02$_{\pm 0.01}$ & 48.7$_{\pm 0.5}$ & \textbf{\underline{47.6$_{\pm 0.3}$}} & 53.8$_{\pm 0.3}$\\
Clamped (C) & 0.76$_{\pm 0.27}$ & \underline{0.09$_{\pm 0.19}$} & \textbf{0.64$_{\pm 0.06}$} & \textbf{\underline{0.13$_{\pm 0.06}$}} & \textbf{0.35$_{\pm 0.07}$} & \textbf{0.10$_{\pm 0.05}$} & \textbf{49.1$_{\pm 1.7}$} & \underline{47.2$_{\pm 0.7}$} & \textbf{\underline{53.9$_{\pm 0.2}$}}\\
Log (L) & \textbf{\underline{0.98$_{\pm 0.02}$}} & \textbf{\underline{0.55$_{\pm 0.26}$}} & 0.63$_{\pm 0.07}$ & 0.06$_{\pm 0.01}$ & 0.31$_{\pm 0.07}$ & 0.05$_{\pm 0.01}$ & 48.1$_{\pm 0.6}$ & 46.8$_{\pm 0.4}$ & 52.9$_{\pm 0.5}$\\
Softcap (S) & 0.75$_{\pm 0.21}$ & \underline{0.05$_{\pm 0.12}$} & 0.58$_{\pm 0.08}$ & 0.10$_{\pm 0.04}$ & 0.31$_{\pm 0.05}$ & 0.07$_{\pm 0.03}$ & 48.5$_{\pm 0.7}$ & \underline{47.4$_{\pm 0.8}$} & 53.7$_{\pm 0.2}$\\
\midrule
C+E & \underline{0.95$_{\pm 0.06}$} & \underline{0.07$_{\pm 0.11}$} & 0.53$_{\pm 0.06}$ & 0.04$_{\pm 0.04}$ & 0.31$_{\pm 0.04}$ & 0.03$_{\pm 0.02}$ & 48.2$_{\pm 0.3}$ & \underline{47.5$_{\pm 0.4}$} & 53.8$_{\pm 0.2}$\\
C+L & \textbf{\underline{0.98$_{\pm 0.02}$}} & \underline{0.55$_{\pm 0.26}$} & 0.63$_{\pm 0.07}$ & 0.06$_{\pm 0.01}$ & 0.31$_{\pm 0.07}$ & 0.05$_{\pm 0.01}$ & 48.1$_{\pm 0.6}$ & 46.8$_{\pm 0.4}$ & 52.9$_{\pm 0.5}$\\
C+S & 0.72$_{\pm 0.20}$ & 0.04$_{\pm 0.09}$ & 0.61$_{\pm 0.02}$ & 0.10$_{\pm 0.04}$ & 0.31$_{\pm 0.04}$ & 0.07$_{\pm 0.03}$ & 48.3$_{\pm 0.5}$ & \underline{47.4$_{\pm 0.4}$} & 53.5$_{\pm 0.2}$\\
E+L & \underline{0.96$_{\pm 0.01}$} & \textbf{\underline{0.80$_{\pm 0.27}$}} & \textbf{\underline{\underline{0.86$_{\pm 0.05}$}}} & \textbf{\underline{\underline{0.67$_{\pm 0.24}$}}} & \textbf{\underline{\underline{0.46$_{\pm 0.11}$}}} & \textbf{\underline{0.42$_{\pm 0.06}$}} & 48.6$_{\pm 0.9}$ & \textbf{\underline{47.6$_{\pm 0.5}$}} & \textbf{\underline{54.0$_{\pm 0.5}$}}\\
E+S & \underline{0.97$_{\pm 0.02}$} & \underline{0.06$_{\pm 0.08}$} & 0.49$_{\pm 0.03}$ & 0.02$_{\pm 0.00}$ & 0.28$_{\pm 0.04}$ & 0.02$_{\pm 0.01}$ & 48.1$_{\pm 0.3}$ & \underline{47.5$_{\pm 0.9}$} & \underline{53.9$_{\pm 0.2}$}\\
L+S & \underline{0.96$_{\pm 0.02}$} & \underline{0.40$_{\pm 0.29}$} & 0.61$_{\pm 0.05}$ & 0.05$_{\pm 0.01}$ & 0.27$_{\pm 0.06}$ & 0.04$_{\pm 0.01}$ & \textbf{48.8$_{\pm 1.9}$} & 46.3$_{\pm 0.6}$ & 53.5$_{\pm 1.1}$\\
\midrule
C+E+L & \underline{0.97$_{\pm 0.01}$} & \underline{0.85$_{\pm 0.25}$} & \textbf{\underline{\underline{0.86$_{\pm 0.05}$}}} & \textbf{\underline{\underline{0.67$_{\pm 0.24}$}}} & \textbf{\underline{\underline{0.46$_{\pm 0.11}$}}} & \textbf{\underline{\underline{0.43$_{\pm 0.05}$}}} & 48.5$_{\pm 0.8}$ & \textbf{\underline{47.7$_{\pm 0.5}$}} & 53.8$_{\pm 0.3}$\\
C+E+S & \underline{0.98$_{\pm 0.03}$} & \underline{0.08$_{\pm 0.11}$} & 0.43$_{\pm 0.04}$ & 0.01$_{\pm 0.01}$ & 0.28$_{\pm 0.04}$ & 0.03$_{\pm 0.01}$ & 48.3$_{\pm 0.6}$ & \underline{47.2$_{\pm 0.3}$} & \textbf{\underline{53.9$_{\pm 0.4}$}}\\
C+L+S & \underline{0.96$_{\pm 0.02}$} & \underline{0.40$_{\pm 0.29}$} & 0.62$_{\pm 0.06}$ & 0.05$_{\pm 0.01}$ & 0.27$_{\pm 0.06}$ & 0.04$_{\pm 0.01}$ & \textbf{48.7$_{\pm 2.0}$} & 46.3$_{\pm 0.6}$ & 53.5$_{\pm 1.1}$\\
E+L+S & \textbf{\underline{\underline{0.99$_{\pm 0.01}$}}} & \textbf{\underline{\underline{0.98$_{\pm 0.01}$}}} & \underline{0.83$_{\pm 0.05}$} & \underline{0.58$_{\pm 0.15}$} & \underline{0.43$_{\pm 0.05}$} & \underline{0.38$_{\pm 0.07}$} & 48.6$_{\pm 0.5}$ & \underline{47.5$_{\pm 0.4}$} & 53.4$_{\pm 0.2}$\\
\midrule
C+E+L+S & \textbf{\underline{0.98$_{\pm 0.01}$}} & \textbf{\underline{\underline{0.98$_{\pm 0.02}$}}} & \textbf{\underline{0.83$_{\pm 0.04}$}} & \textbf{\underline{0.57$_{\pm 0.15}$}} & \textbf{\underline{0.44$_{\pm 0.06}$}} & \textbf{\underline{0.38$_{\pm 0.06}$}} & \textbf{\underline{\underline{49.5$_{\pm 2.0}$}}} & \textbf{\underline{47.4$_{\pm 0.5}$}} & \textbf{53.4$_{\pm 0.1}$}\\
\bottomrule
\end{tblr}

\caption{
\label{table:results-decoders}
Area under the PK, NIAH$_s$, and NIAH$_m$ curves (split into in-context and out-of-context), and decoder benchmark results for both training experiments averaged over five runs.
Best per-group result in bold; results equal to or better than the ALiBi baseline are underlined; best result per column is double underlined.
}
\end{table*}

\subsection{Training to Investigate the Failure Mode}
\label{sec:training-failure-mode}

To study attention blindness independently from out-of-context degradation, we train several small causal language models where slopes are set to decrease/increase the expected blindness distance.

\paragraph{Model Architecture and Data} We investigate Llama~\cite{touvron:2023} decoder models following the SmolLM~\cite{ben-allal:2024} recipe.
Training data consists of 20B FineWeb-Edu tokens, disjoint from the perplexity test set.

\paragraph{Training and Evaluation} We train for one epoch with AdamW~\cite{loshchilov:2019}, not aiming for state-of-the-art performance, but for models sufficiently trained to compare ALiBi variations. 
We repeat each configuration over five seeds, but fix the dataset order so that all remaining variation is due to positional encoding (which is the variable under study) and model initialization (to account for variation).
Full hyperparameters are listed in Appendix Section~\ref{app:training-additional-details}.

We perform zero-shot evaluation on standard decoder benchmarks that are suitable for 148M models,\hspace{-0.2em}\footnote{See Appendix Section~\ref{app:benchmarks} for a detailed list.} as well as on PK and NIAH. Benchmarks are grouped into common sense (CS), question answering (QA), and language (LG).

\paragraph{Positional Encoding Configurations} We devised several configurations varying the expected blindness distances:
\emph{\Ni~Steep:}~A stress test with steep slopes ($\delta_1 = 128$, $\delta_H = 512$) encouraging early attention blindness.
\emph{\Nii~Safe:} The counterposition to the first test, where little to no blindness should occur ($\delta_1 = 2{,}048$, $\delta_H = 4{,}096$).
\emph{\Niii~Wide:}~This distributes slopes so that $\delta_1, ...,\delta_H$ cover a wide range of distances in powers of two ($\delta_1 = 16$, $\delta_H = 4{,}096$).
Finally, we also add ALiBi and RoPE baselines for comparison.

\subsubsection{Results}

Results are shown in Table~\ref{table:results-decoders}a. We report the mean and standard deviation over five runs.
Steep slopes achieve the lowest scores overall.
Safe outperforms the ALiBi baseline on passkey retrieval by 18 pp within context and by 3 pp out of context, but achieves considerably lower scores for both single and multi-key NIAH. It is competitive on QA and Language, but slightly inferior on CS.

\subsubsection{Interim Discussion} 

As expected, the steep variant performs worst on NIAH among ALiBi strategies, but contrary to our expectation it does not fail completely but is even strong on out-of-context PK.
Safe is strongest or competitive on PK, but this advantage seems not to carry over to NIAH, where it falls behind the ALiBi baseline (0.46 vs 0.65 in context, 0.01 vs 0.13 out-of-context).
Wide is mediocre on PK retrieval, weak on both NIAH tasks, but performs strongest on QA, and strong on LG tasks.

The RoPE baseline achieves competitive scores on the decoder benchmarks but  underperforms on token retrieval, especially out of context.
We hypothesize that our small-model setup with only 20B tokens is favorable for ALiBi's recency bias~\cite{press:2022}.
Still, RoPE is competitive on downstream benchmarks, and it helps to confirm that the ALiBi benchmark scores are within a reasonable range.
Overall, ALiBi proves to be a surprisingly strong baseline on decoder benchmarks and NIAH.

\subsection{Investigating Mitigation Strategies}
\label{sec:training-mitigations}

Finally, we investigate the training-level underflow mitigation strategies presented in Section~\ref{sec:fixes}.

\paragraph{Setup and Mitigation Strategies} The setup is identical to the previous experiment regarding model architecture, training, and evaluation, except that one or more mitigation strategies are enabled (which we abbreviate by their first letter): \Ni~For the clamping  strategy \emph{(C)}, we use $c_{clamp} = -87$, which is a defensive choice to not only avoid underflow but also the subnormal\footnote{Thresholds and additional details in Appendix Section~\ref{app:floating-point-thresholds}.} floating point numbers~\cite{goldberg:1991}. 
\Nii~For the explicit slopes \emph{(E)}, we use the recipe proposed in Section~\ref{sec:fixes} but targeted at four times the training context length ($\delta_H = 8{,}192$).
\Niii~For soft capping \emph{(S)}, we follow \citet{riviere:2024} and adopt $z=50$.
\Niv~Log distance \emph{(L)} is applied as described in \Cref{sec:fixes}. E and L interact, since E assumes linear scaling. We therefore adapt E's slopes to log-scaled distances with $m_h = 9.2 / \log(\delta_h + 1)$ whenever active together, elaborated in Appendix Section~\ref{app:details}.

\subsubsection{Results}

The results are presented in Table~\ref{table:results-decoders}b, grouped by the number of combined mitigations.
In contrast to ALiBi (Table~\ref{table:results-decoders}a), quite a few mitigations improve PK or NIAH scores---some by a considerable margin.
The decoder benchmarks, on the other hand, show only minor differences and therefore we focus on associative retrieval in the following.

Multiple mitigation strategies achieve small or even substantial improvements over the ALiBi baseline on associative retrieval:
On PK, ALiBi performs weakest and several strategies  considerably outperform it in both in- and out-of-context performance.
On NIAH$_s$ and NIAH$_m$, however, ALiBi achieves strong results that are only surpassed by four strategies.
Notably, the combination E+L is part of all of the strongest strategies.
Some configurations show a large standard deviation, likely due to our experiments with small models, so their results should be interpreted with caution.

\subsubsection{Interim Discussion}

Most notably, clamping does not affect the results substantially, suggesting that underflow does not necessarily have an effect on model performance. 

The L strategy shows high effectiveness on PK, but its NIAH scores are slightly below ALiBi in context and considerably below it out of context. Its NIAH scores increase when the slopes are also adjusted for log scale (E+L). Applying multiple mitigations in combination mostly leads to adverse effects, widening the gap to ALiBi. The only exception is combinations including E+L, all of which outperform the ALiBi baseline. S seems to deteriorate token retrieval results unless when E+L is involved, where it contributes to the highest out-of-context PK scores. We adopted the soft cap interval $z$ from \citet{riviere:2024}, which is in hindsight likely too large for linear scaling, requiring log scaling to be effective.

Contrary to our expectations, the E~strategy underperforms when used in isolation. While it is strong on PK, it is weaker than ALiBi at both NIAH tasks, and most importantly, it degrades out-of-context performance for all token retrieval tasks. Its slopes are tailored to four times the training context length, covering beyond the maximum length in the evaluation.
We conclude that avoiding attention blindness alone does not suffice: the very flat-sloped heads act as retrieval heads~\cite{wu:2025}, whose weak positional signal makes them useful not only far beyond $\delta_H$, but already at short to medium range, long before they go blind.
The Safe strategy from \Cref{sec:training-failure-mode} acts as an important complement: It lacks steep slopes (local heads), which results in high in-context performance on PK, but deteriorates all NIAH scores compared to the baseline. Both very steep and very flat slopes appear to be required for successful token retrieval\mbox{.\hspace{-0.2em}}%
\footnote{We performed additional head ablations (Appendix Section~\ref{app:head-ablation}), which support the hypothesis that both very steep and flat heads are required for retrieval.}

\section{Synthesis and Discussion}

Having confirmed that underflow occurs in pretrained models, and having varied its expected onset in our training experiments, we now turn to what this means for model behavior and practice.

\subsection{On the Underflow}

\emph{Why has this gone unnoticed for so long?} As our experiments show (Section~\ref{sec:training-mitigations}), underflow can be a silent failure: tiny, almost-zero attention scores become exact zeros, which can easily be overlooked.

\emph{Is underflow a problem?} At the scale we study, not measurably: clamping leaves all metrics unchanged (Section~\ref{sec:training-mitigations}). The presented problem, however, affects all previous models trained with ALiBi, including ALiBi variants and other positional encodings that operate through linear-scaling additive biases~(e.g., TISA~\cite{wennberg:2021}). 
At minimum it is a computational inefficiency: unlike an explicit sliding window, the underflowing entries are still computed before being discarded, wasting computation time on token pairs that will never contribute to the output.

Beyond this forward-pass inefficiency, underflow may matter most during training. ALiBi already drives distant token weights toward near-zero, so during the forward pass, an underflowed exact zero and a small nonzero weight both contribute little. They, however, differ in the backward pass, where only the nonzero weight still carries a small gradient. If this account is correct, the practical stakes of underflow lie less in inference-time robustness than in whether steep-sloped heads can ever learn to use information beyond their own blindness distance. Since our experiments train models from scratch, this distinction is present in every run. We leave this hypothesis for future work.

\subsection{Implications for ALiBi}

Our experiments confirm that ALiBi's default slopes are a strong baseline. However, they also show that ALiBi's extrapolation capabilities are limited, which previous works already observed~\cite{kazemnejad:2023,chi:2023, gelberg:2026}. More specifically, we have also shown that underflow is easy to overlook, at least with earlier decoder benchmarks for smaller models, which, however, could be an inherent limitation of the benchmarks. Under comparable decoder benchmark results, we observe substantial improvements on the associative retrieval tasks, suggesting that the validation of models, including but likely not limited to ALiBi, should closely monitor associative retrieval.

While most ALiBi models use the empirical standard slope recipe by \citet{press:2022} (as shown in Appendix Table~\ref{table:models}), which our experiments confirmed to be strong, it was never theoretically motivated. Our formalization of the expected blindness distance $\delta_h$ provides an explanation of why the standard slopes are a strong baseline. It was also used to obtain matching slopes for the log-scaled mitigations, which overall excelled in token retrieval.

Extrapolation capabilities are affected by the choice of slopes as evidenced by the E+L combination. 
Tokens, however, can draw on information beyond a single per-layer-and-head's window, increasing the effective window size by the size of the local window through every layer~\cite{chi:2023}.
Thereby information reaches tokens indirectly, through several layers of relay, but likely at a loss of information~\cite{dong:2021b}.
Through the steep variant we have learned that even suboptimal slopes can recover part of the performance.
Suspecting this indirect information flow, we performed a sensitivity analysis over model depth (Appendix Table~\ref{table:depth-analysis}). With the SmolLM recipe, we started with a deep model, but reducing the model depth reduces both in- and out-of-context performance.
This suggests that the indirect information flow dominates and is responsible for the models recovering performance despite being partially blind.

ALiBi is effectively a sliding window over the attention matrix~\cite{beltagy:2020} as previously recognized by \citet{chi:2023} and \citet{oka:2025}. 
The difference is that it also inversely scales with large token logits, thereby also depending on the data.
Recall that a considerable fraction of weights is unusable due to attention blindness. For a steep head with an effective window of $124$ tokens and a $2{,}048$-token context, 88.3\% of causal token pairs are beyond that window. A fixed sliding window would allow computational optimizations to ignore out-of-window attention weights instead of processing and then discarding them.

\subsection{Recommendations for Model Training}

As we only trained small models, we outline how we would train larger models.

As a replacement for vanilla ALiBi, we recommend log scaling with explicit slopes (E+L), which seems to be highly beneficial for out-of-context extrapolation in associative retrieval.
The slopes must be adapted when using log-scaled distances: log scaling with default slopes (L) improves PK but falls behind ALiBi on NIAH.

Clamping was neutral in our experiments, but is a cheap safeguard that preserves ALiBi's soft sliding window without the hard cutoff caused by underflow.
As an alternative, a hard sliding window can be applied to ALiBi~\cite{chi:2023,jelassi:2024} for which attention computation could be optimized to ignore the out-of-window attention entries, thereby improving efficiency.

Based on the observed results, we cannot recommend soft capping as is, but restricting the range of attention logits should be beneficial since (a)~this can also cause the underflow and (b)~large logit outliers might not be affected that much by a smaller linear-scaled bias.
There are other means to achieve a similar effect, such as nGPT~\cite{loshchilov:2025} which extends the normalization in the architecture, which might be an alternative.

\section{Conclusions}

This paper identified a failure mode present in all previous models with the ALiBi positional encoding.
We analytically showed the occurrence of underflow and blind attention heads, and empirically confirmed the issue in pre-trained models.
We trained 148M-parameter decoder models, including a series of isolated and combined mitigation strategies.
The identified deficiency does not necessarily result in a defective model: even configurations designed to increase blindness remained within acceptable range on standard benchmarks.
The proposed mitigations showed only minor differences (of $-1.0$ to $0.8$~pp) on downstream benchmarks (CS, QA, LG), but considerable improvements associative token retrieval (of up to 19/93~pp on PK, 21/54~pp on NIAH$_s$, and 10/32~pp on NIAH$_m$, in/out of context). Overall, we conclude that underflow is a silent failure which, at least at the scale we study, does not by itself impair ALiBi models, whereas the choice of slopes and distance scaling matters considerably, and is best exposed by associative retrieval rather than standard benchmarks.

\clearpage
\section*{Limitations}

To balance computational costs with the need for rigorous, wide-ranging empirical experimentation, our experiments were conducted with 148M-parameter decoder models.
While the eventual underflow and the resulting attention blindness are analytically grounded, not all observations and findings may translate to larger models.

Furthermore, we held architecture and training corpus fixed to isolate the effect of the ALiBi variations and mitigation strategies.
As a result, we do not explore how these findings interact with different architectural choices or corpora.
We leave a systematic, considerably more compute-intensive, hyperparameter sweep of the most promising configurations for future work.

\section*{Ethical Considerations}

This work identifies a numerical failure mode in an existing positional encoding and evaluates it on well-known publicly released pretrained models (BLOOM, Falcon-RW, MPT) through inference-time evaluation.
We do not fine-tune or redistribute any of those models' weights.
Our own model training experiments use FineWeb-Edu, a public filtered web corpus.
The PK retrieval task consists entirely of template text and NIAH uses parts of FineWeb-Edu as well.
Neither our training nor evaluation data introduces personal or sensitive information beyond what is already in the corpus.

Training the 148M-parameter decoder models as described in Sections~\ref{sec:training-failure-mode} and \ref{sec:training-mitigations} required non-trivial compute.
We constrained model size and training budget to make this feasible as well as keep the costs justifiable.
We release our code so that others can adopt our experimental setup rather than reproduce it from scratch.

The identified failure mode is inherent to the ALiBi positional encoding and depends primarily on token distance rather than on specific inputs. We therefore do not see potential for misuse beyond the well-known limitation that ALiBi models attend less to distant tokens.
We do not foresee direct negative societal impacts from this work beyond those generally associated with language models.
\section*{Acknowledgments}

We sincerely thank Lalith Manjunath, Stefan Schweter, and Wilhelm Pertsch for fruitful discussions on ALiBi and positional encoding, and Daniel Gallagher for feedback on the manuscript. We thank Claude-Pierre Jeannerod for bringing relevant related work to our attention. We thank the anonymous reviewers for the constructive and thorough reviews, which helped to improve this paper.

This work was done as part of the \href{https://coral-nlp.github.io}{CORAL project} which was funded by the German Federal Ministry of Research, Technology, and Space (BMFTR) under the grant number 16IS24077A and 16IS24077B.
The authors gratefully acknowledge the computing time made available to them on the high-performance computer at the NHR~Center of TU~Dresden. This center is jointly supported by the Federal Ministry of Research, Technology and Space of Germany and the state governments participating in the NHR.

The authors used AI assistance for language editing and coding. 
For the manuscript, GLM-5.2-FP8 and Claude were used to identify stylistic and grammatical issues. The models never made any changes to the document.
For the experiments, Claude Code was used, with all contributions critically reviewed by the authors. The authors take full responsibility for all content presented in this paper.

\bibliography{arr26-alibi-deficiencies-lit}
\appendix

\section{Floating Point Thresholds}

\label{app:floating-point-thresholds}

In Table~\ref{table:floating-point-thresholds}, we list the floating point thresholds for PyTorch in a Linux 64-bit system.

\begin{table}[!hptb]%
\centering%
\small
\SetTblrInner{rowsep=1pt,colsep=8pt}
\begin{tblr}[expand=\cellhl]{
  colspec={@{}lclS[table-format=3.4]@{}},
} 
\toprule
& & \SetCell[c=2]{c} \textbf{Formats}\\
\cmidrule[l]{3-4} \textbf{Name} & \textbf{Condition} & Format & \SetCell[c=1]{r} {Value}\\
\midrule
Underflow $\tau_{u}$ & $x \leq \tau_{u}$ & bf16 & -92.1886 \\
&  & fp32 & -103.2789\\
Denorm $\tau_{d}$ & $ \tau_{u} < x \leq \tau_{d}$ & bf16 & -87.3365\\
&  & fp32 & -87.3365\\
Overflow $\tau_{o}$ & $\tau_{d} \leq x$ & bf16 & 88.7189\\
&  & fp32 & 88.7228\\
\bottomrule
\end{tblr}
\caption{Thresholds for underflow, denormalization, and overflow each in both 32-bit full precision (fp32) and brain float 16-bit half-precision (bf16).}
\label{table:floating-point-thresholds}
\end{table}
\begin{table*}[!bth]%
\centering%
\small
\SetTblrInner{rowsep=1pt,colsep=4pt}
\begin{tblr}[expand=\cellhl]{
  colspec={@{}p{4cm}@{}llrlcrrr@{}},
  row{1}  = {font=\bfseries},
} 
\toprule
Model/Model Family & Arch. & Type & Size(s) & Language(s) & Public & Window & Heads & Slopes\\
\midrule
Baichuan-13B-Base$^1$ & dec & base & 13B & zho, eng & y & 4,096 & 40 & default\\
Baichuan-13B-Chat & dec & chat & 13B & zho, eng & y & 4,096 & 40 & default\\
BloombergGPT$^2$ & dec & base & 50B & eng & n & 2,048 & 40 & unspecified$^\dagger$\\
bigscience/bloom-560m$^3$ & dec & base & 560M & 46 lang., code & y & 2,048 & 16 & default\\
bigscience/bloom-1.1b$^3$ & dec & base & 1.1B & 46 lang., code & y & 2,048 & 16 & default\\
bigscience/bloom-1.7b$^3$ & dec & base & 1.7B & 46 lang., code & y & 2,048 & 16 & default\\
bigscience/bloom-3b$^3$ & dec & base & 3B & 46 lang., code & y & 2,048 & 32 & default\\
bigscience/bloom-7.1b$^3$ & dec & base & 7B & 46 lang., code & y & 2,048 & 32 & default\\
bigscience/bloom$^3$ & dec & base & 176B & 46 lang., code & y & 2,048 & 112 & default\\
BTLM-3B-8k-base$^4$ & dec & base & 3B & eng & y & 8,192 & 32 & default\\
BTLM-3B-8k-chat$^4$ & dec & chat & 3B & eng & y & 8,192 & 32 & default\\
CodeLLM$^5$ & dec & base & 1B & code & y & 1,024 & 32 & default\\
Falcon-RW 1B$^6$ & dec & base & 1B & eng & y & 2,048 & 32 & default\\
Falcon-RW 7B$^6$ & dec & base, chat & 7B & eng & y & 2,048 & 64 & default\\
jais-family-590m$^7$ & dec & base, chat & 590M & arb, eng & y & 2,048 & 12 & default\\
jais-family-1p3b$^7$ & dec & base, chat & 1.3B & arb, eng & y & 2,048 & 16 & default\\
jais-family-2p7b$^7$ & dec & base, chat & 2.7B & arb, eng & y & 2,048 & 20 & default\\
jais-family-6p7b$^7$ & dec & base, chat & 6.7B & arb, eng & y & 2,048 & 32 & default\\
jais-family-13b$^7$  & dec & base, chat & 13B & arb, eng & y & 2,048 & 40 & default\\
jais-family-30b-8k$^7$  & dec & base & 30B & arb, eng & y & 8,192 & 56 & default\\
jais-family-30b-16k$^7$  & dec & base & 30B & arb, eng & y & 16,384 & 56 & default\\
MPT 7B$^8$ & dec & base, chat, inst. & 7B & eng & y & 2,048 & 32 & default\\
MPT 7B StoryWriter$^8$ & dec & fine-tuned & 7B & eng & y & 65,536 & 32 & default\\
jina-embeddings-v2-base-en$^9$ & enc & embeddings & 137M & eng & y & 8,192 & 12 & default\\
MosaicBERT-base$^{10}$ & enc & base & 137M & eng  & y & 128 & 12 & default\\
MosaicBERT-256$^{10}$ & enc & base & 137M & eng & y & 256 & 12 & default\\
MosaicBERT-512$^{10}$ & enc & base & 137M & eng & y & 512 & 12 & default\\
MosaicBERT-1024$^{10}$ & enc & base & 137M & eng & y & 1,024 & 12 & default\\
MosaicBERT-2048$^{10}$ & enc & base & 137M & eng & y & 2,048 & 12 & default\\
\bottomrule
\end{tblr}
\caption{Overview of models trained with the ALiBi positional encoding. 
$^1$\citet{yang:2025}, $^2$\citet{wu:2023}, $^3$\citet{lescao:2022}, $^4$\citet{dey:2023}, $^5$\citet{kazemnejad:2023}, $^6$\citet{penedo:2023}, $^7$\citet{sengupta:2023}, $^8$\citet{mosaic:2023}, $^9$\citet{guenther:2024}, $^{10}$\citet{portes:2023}. $\dagger$:~Slopes were unspecified in the corresponding publication but since the BLOOM recipe was closely followed for training, it likely uses default slopes.}
\label{table:models}
\end{table*}

\section{Softmax}

\subsection{Stable Softmax}
\label{app:stable-softmax}
In computational applications, stable softmax, a safer formulation of softmax is often used. It first subtracts the maximum element $s = \max x_i$ from every other element~\cite{goodfellow:2016}:

$$\mathrm{softmax}\!\left(\mathrm{x}\right) = \frac{e^{x_i - s}}{\sum_{j=1}^{k}e^{x_j - s}}$$

\noindent This shift, however, only protects from {\em over}- but not underflow.

\subsection{Softmax Invariance}
\label{app:softmax-invariance}

Softmax is translation invariant which means:

$$\mathrm{softmax}\!\left(\mathrm{x} + c\right) = \mathrm{softmax}\!\left(\mathrm{x}\right).$$

\noindent Some model implementations such as BLOOM's Hugging Face implementation use this property to transform the negative ALiBi bias $-m(i-j)$ into a positive-slope term:
\begin{align*}
\mathrm{softmax}\!\left(A + B\right) &= \mathrm{softmax}\!\left(A - m(i-j)\right)\\
&= \mathrm{softmax}\!\left(A - mi + mj\right)\\
&= \mathrm{softmax}\!\left(A + mj\right)
\end{align*}

\noindent 
Due to the aforementioned stable softmax, the resulting term will still underflow before overflowing.
For readability and consistency we stick to the underflow framing which was also used in the original ALiBi paper \cite{press:2022}.

\section{Implementation and Reproduction}

Since the widely-used FlashAttention~\cite{dao:2022,dao:2024} did not support clamping at the bias level, log scaling, and soft capping, we implemented a solution based on PyTorch's FlexAttention.
Correctness was ensured through preliminary tests, in which we directly compared against FlashAttention, where \mbox{Flash-/FlexAttention} was the only differing variable.

\subsection{Implementations of Existing Models}

BLOOM and Falcon add the bias instead of subtracting it, using the shift-invariant $+mj$ form (\Cref{app:softmax-invariance}).
While this looks like it would shift the failure toward overflow rather than underflow, the stable softmax (\Cref{app:stable-softmax}) converts it back to the same underflow behavior analyzed in the main text.

\subsection{Environment}

All experiments were conducted on a Linux system with CUDA 12.8 and Python 3.12 installed.
Each configuration was executed on a single NVIDIA H100 with 94 GiB RAM.
Training and evaluation runs, including development and failed attempts, amounted to approximately $9{,}000$~GPUh.

\section{Details on Approach and Models}
\label{app:details}

\subsection{Mitigation Strategies}

For the E+L combination, we adapted the slopes to $m_h = 9.2 / \log(\delta_h + 1)$. 
The value $9.2$ is the value of $\log(10{,}000)$, which is a heuristic that controls the numeric range of the resulting slopes, and thereby ALiBi biases.
Instead of blindness distances, we adjusted the slopes for distances of $32, 64, \dots, 8192$.
Throughout this work $\log$ denotes the natural logarithm.

\subsection{Pre-Trained Models}
\label{app:models}

In Appendix Table~\ref{table:models}, we compiled a list of models which, to the best of our knowledge, covers all {{ALiBi}} models that are accompanied by a publication, plus all models we could identify on Hugging~Face~Hub.

\subsection{Blindness Distances for Trained Models}

In Table~\ref{table:exp2-slopes} we list the expected blindness distances for all trained model configs.
\begin{table*}[!hptb]%
\small
\centering%
\SetTblrInner{rowsep=1pt,colsep=4pt}
\begin{tblr}[expand=\cellhl]{
  colspec={@{}lrrrrrrrrr@{}},
  row{1}  = {font=\bfseries},
} 
\toprule
\textbf{Name} & \bm{$\delta_1$} & \bm{$\delta_2$} & \bm{$\delta_3$} & \bm{$\delta_4$} & \bm{$\delta_5$} & \bm{$\delta_6$} & \bm{$\delta_7$} & \bm{$\delta_8$} & \bm{$\delta_9$}\\
\midrule
Steep & 128 & 152 & 181 & 215 & 256 & 304 & 362 & 431 & 512\\
Safe & 2,048 & 2,233 & 2,435 & 2,656 & 2,896 & 3,158 & 3,444 & 3,756 & 4,096\\
Wide & 16 & 32 & 64 & 128 & 256 & 512 & 1,024 & 2,048 & 4,096\\
ALiBi (default) & 124 & 175 & 349 & 699 & 1,397 & 2,795 & 5,590 & 11,179 & 22,358\\
\midrule
Explicit & 32 & 64 & 128 & 256 & 512 & 1,024 & 2,048 & 4,096 & 8,192\\
Log & * & * & * & * & ** & ** & ** & ** & **\\
\midrule
C+L & * & * & * & * & ** & ** & ** & ** & **\\
C+E & 32 & 64 & 128 & 256 & 512 & 1,024 & 2,048 & 4,096 & 8,192\\
E+L & * & * & * & * & * & * & * & * & *\\
E+S & 32 & 64 & 128 & 256 & 512 & 1,024 & 2,048 & 4,096 & 8,192\\
L+S & * & * & * & * & ** & ** & ** & ** & **\\
\midrule
C+E+L & * & * & * & * & * & * & * & * & *\\
C+E+S & 32 & 64 & 128 & 256 & 512 & 1,024 & 2,048 & 4,096 & 8,192\\
C+L+S & * & * & * & * & ** & ** & ** & ** & **\\
E+L+S & * & * & * & * & * & * & * & * & *\\
\midrule
C+E+L+S & * & * & * & * & * & * & * & * & *\\
\bottomrule
\end{tblr}
\caption{Expected blindness distances (rounded to the nearest integer) for all ALiBi models trained in this work. Distances are computed with a threshold of $\tau_d \approx -87.34$, which is identical for bf16 and fp32. Mitigations or combinations thereof that were omitted use default slopes. $*$: $\delta_x>10^9$, $**$: $\delta_x\gg10^9$.}
\label{table:exp2-slopes}
\end{table*}

\subsection{Why Default Slopes Are a Strong Baseline}

With nine heads, the default slopes have blindness distances from $124$ to $22{,}358$ tokens (Table~\ref{table:exp2-slopes}), i.e., they contain both strongly local and effectively global heads.
Our results indicate that both ends of this range are required: removing either the steepest or the flattest heads (nearly) eliminates retrieval (Table~\ref{table:head-ablations}).
Moreover, the Safe configuration, which was expected to perform well, lacks steep heads and falls behind the baseline on NIAH.

\begin{table*}[!hbpt]%
\small
\centering%
\SetTblrInner{rowsep=1pt,colsep=4pt}
\begin{tblr}[expand=\cellhl]{
  colspec={@{}lp{10cm}p{3cm}@{}},
  row{1}  = {font=\bfseries},
} 
\toprule
Name &  Description & Publication\\
\midrule
\SetCell[c=3]{c} \textbf{Common Sense}\\
\midrule
HellaSwag & Given a description, choose the most likely followup. & \cite{zellers:2019}\\
PIQA & Given a physical goal, choose the most sensible solution. & \cite{bisk:2020}\\
WinoGrande & Common sense reasoning on pronoun resolution problems. & \cite{sakaguchi:2021}\\
COPA & Given a premise, choose the more plausible out of two alternatives. & \cite{roemmele:2011}\\
SiQA & Multiple choice question on commonsense reasoning about social interactions. & \cite{sap:2019}\\
WSC & Identify one of two sentences as correct through referential ambiguity. & \cite{levesque:2012}\\
\midrule
\SetCell[c=3]{c} \textbf{Question Answering}\\
\midrule
ARC Easy & Advanced question answering. Easy subset. & \cite{clark:2018}\\
ARC Challenge & Advanced question answering. Challenging subset. & \cite{clark:2018}\\
BoolQ & Yes/no question answering to test reading comprehension. & \cite{clark:2019}\\
SciQ & Multiple choice science exam questions. & \cite{welbl:2017}\\
RACE & Reading comprehension drawn from English-as-a-foreign-language exams. & \cite{lai:2017}\\
\midrule
\SetCell[c=3]{c} \textbf{Language}\\
\midrule
LAMBADA & Tests text understanding through word prediction requiring broad discourse context. & \cite{paperno:2016}\\
BLiMP & Given a minimal pair of sentences, predict which one is grammatically acceptable. & \cite{warstadt:2020}\\
WiC & Given a target word and two sentences, predict whether it is used with the same meaning. & \cite{pilehvar:2019}\\
\bottomrule
\end{tblr}
\caption{List of (grouped) decoder benchmarks used in the experiments.}
\label{table:benchmarks}
\end{table*}

\section{Experimental Design}

\subsection{Associative Retrieval Probes}

Following \citet{mohtashami:2023} the PK value is stated twice; following \citet{kamradt:2023} the NIAH value is stated once. We adhered to the original formulations, but this asymmetry should be kept in mind when interpreting the results.

\subsubsection{Passkey Retrieval}
\label{app:passkey-retrieval}

The following prompt was used:

\begin{quote}
\small\ttfamily
The pass key is \{pk\}. The pass key is \{pk\}.
[The grass is green. The sky is blue. The sun is yellow. Here we go. There and back again.]*
The pass key is
\end{quote}

\noindent The filler text was adopted from \citet{zhu:2024}.

\subsubsection{Needle in a Haystack}

The needle is embedded in existing documents as follows:

\begin{quote}
\small\ttfamily
[Content before.]
The secret code for today is \{pk\}.
[Content after.] The secret code for today is
\end{quote}

\noindent For the multi-key variant:
~
\begin{quote}
\small\ttfamily
[Content before.]
The secret code for key: \{key\} is value: \{value\}.
[Content.] The secret code for key: \{key\} is value: \{value\}.
[Content.]
The secret code for key: \{key\} is value: \{value\}.
[Content after.] The secret code for key: \{key\} is value:
\end{quote}

\subsection{Benchmarks}
\label{app:benchmarks}

In the experiments we relied on well-known existing decoder benchmarks that we summarize in Table~\ref{table:benchmarks}.
For evaluating against the benchmarks we used \href{https://github.com/EleutherAI/lm-evaluation-harness}{lm-evaluation-harness} in a zero-shot setting.

\begin{table}[!htbp]%
\centering%
\small
\SetTblrInner{rowsep=1pt,colsep=4pt}
\begin{tblr}[expand=\cellhl]{
  colspec={@{}ll@{}},
  row{1}  = {font=\bfseries},
  row{2}  = {font=\bfseries},
  row{12}  = {font=\bfseries},
  row{24}  = {font=\bfseries},
} 
\toprule
Parameter & Value\\
\midrule
\SetCell[c=2]{c} Model Architecture\\
\midrule
Layers & 30\\
Hidden size & 576\\
Attention Type & multi-head\\
Attention Heads & 9\\
Head dimension & 64\\
MLP intermediate size & 1,536\\
Activation & SiLU\\
Norm & RMSNorm\\
Context length & 2,048\\
\midrule
\SetCell[c=2]{c} Training\\
\midrule
Precision & bf16\\
Optimizer & AdamW, $\beta=(0.9, 0.98)$,\\
& $\varepsilon=\num{1e-10}$, wd=\num{1e-5}\\
Learning rate & \num{8e-4}\\
Learning rate schedule & warmup-stable-decay (10 / 80 / 10)\\
Dropout & 0.0\\
Gradient clipping & 1.0\\
Batch size & 48\\
Gradient accumulation & 16\\
AttentionImplementation & FlashAttention (RoPE),\\
 & FlexAttention (others)\\
\midrule
\SetCell[c=2]{c} Tokenizer\\
\midrule
Vocabulary size & 49,152\\
\bottomrule
\end{tblr}
\caption{Hyperparameters used for model training.}
\label{table:hyperparameters}
\end{table}

\begin{table}
\small
\centering
\SetTblrInner{rowsep=1pt,colsep=1pt}
{
\small
\begin{tblr}{
  colspec   = {@{}lrr@{}},
  rowsep    = 2pt,
  colsep    = 6pt,
  row{1}    = {font=\bfseries},
}
\toprule
Name & Train & Valid. \\
\midrule
Steep & 3.085$_{\pm 0.020}$ & 2.988$_{\pm 0.021}$\\
Safe & 2.984$_{\pm 0.002}$ & 2.876$_{\pm 0.000}$\\
Wide & \textbf{\underline{2.904$_{\pm 0.118}$}} & 2.852$_{\pm 0.001}$\\
ALiBi & 2.951$_{\pm 0.002}$ & 2.846$_{\pm 0.001}$\\
RoPE & \underline{2.950$_{\pm 0.003}$} & \textbf{\underline{2.842$_{\pm 0.002}$}}\\
\midrule
Explicit (E) & \textbf{\underline{2.950$_{\pm 0.002}$}} & \textbf{\underline{2.846$_{\pm 0.001}$}}\\
Clamped (C) & \underline{2.951$_{\pm 0.002}$} & 2.846$_{\pm 0.001}$\\
Log (L) & 2.991$_{\pm 0.002}$ & 2.883$_{\pm 0.001}$\\
Softcap (S) & 2.952$_{\pm 0.002}$ & 2.846$_{\pm 0.001}$\\
\midrule
C+E & \textbf{\underline{2.950$_{\pm 0.003}$}} & \underline{2.846$_{\pm 0.001}$}\\
C+L & 2.991$_{\pm 0.002}$ & 2.883$_{\pm 0.001}$\\
C+S & 2.954$_{\pm 0.001}$ & 2.846$_{\pm 0.000}$\\
E+L & 2.961$_{\pm 0.002}$ & 2.856$_{\pm 0.001}$\\
E+S & 2.952$_{\pm 0.003}$ & \textbf{\underline{2.846$_{\pm 0.000}$}}\\
L+S & 2.991$_{\pm 0.003}$ & 2.883$_{\pm 0.002}$\\
\midrule
C+E+L & 2.961$_{\pm 0.002}$ & 2.856$_{\pm 0.001}$\\
C+E+S & \textbf{\underline{2.951$_{\pm 0.002}$}} & \textbf{\underline{2.846$_{\pm 0.000}$}}\\
C+L+S & 2.991$_{\pm 0.003}$ & 2.883$_{\pm 0.002}$\\
E+L+S & 2.961$_{\pm 0.003}$ & 2.856$_{\pm 0.001}$\\
\midrule
C+E+L+S & \textbf{2.961$_{\pm 0.003}$} & \textbf{2.856$_{\pm 0.002}$}\\
\bottomrule
\end{tblr}
}
\caption{
\label{table:loss}
Average training and validation loss for all configurations.
}
\end{table}

\section{Training Details and Additional Results}
\label{app:training-additional-details}

Hyperparameters for training and model architecture are listed in Table~\ref{table:hyperparameters}.
Encountered loss values, shown in Table~\ref{table:loss}, were unsurprising and showed little difference.

\subsection{Associative Retrieval Curves}
\label{app:retrieval-curves}

\begin{figure*}[!t]
\centering
\includegraphics{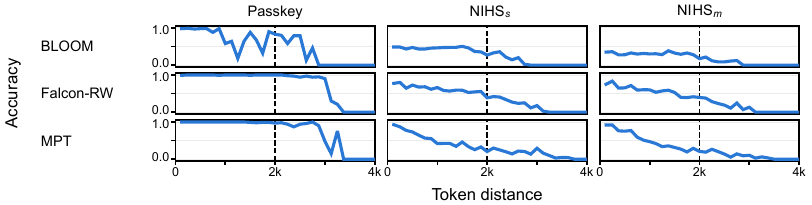}
\caption{Passkey retrieval and NIAH curves for the pre-trained models.}
\label{fig:retrieval-curves-exp1}
\end{figure*}

\begin{figure*}[!t]
\centering
\includegraphics{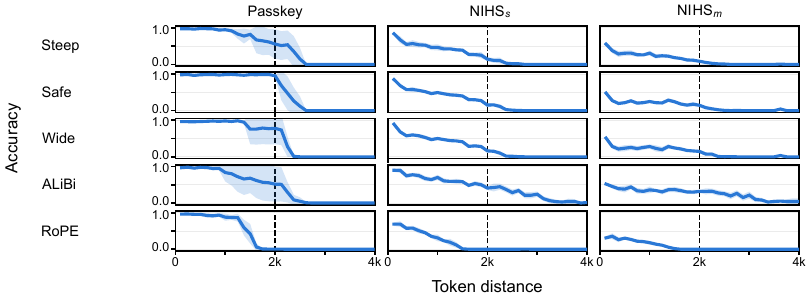}
\caption{Passkey retrieval and NIAH curves for all trained models of the failure mode experiment.}
\label{fig:retrieval-curves-exp2}
\end{figure*}

\begin{figure*}[!t]
\centering
\includegraphics{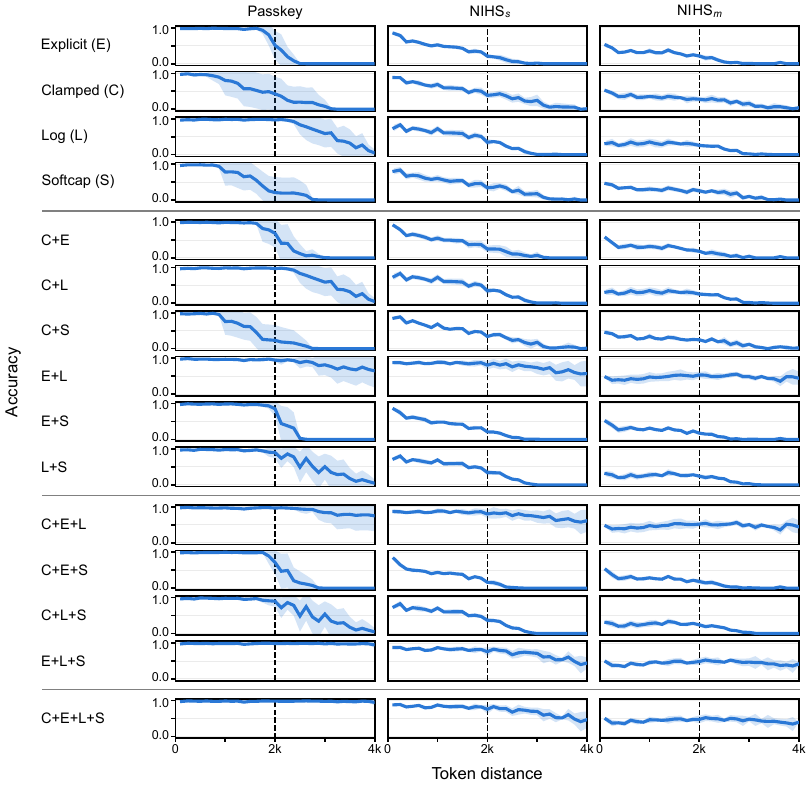}
\caption{Passkey retrieval and NIAH curves for all trained mitigation strategies.}
\label{fig:retrieval-curves-exp3}
\end{figure*}

PK and NIAH curves, averaged over five runs, are shown in Figures~\ref{fig:retrieval-curves-exp1}, \ref{fig:retrieval-curves-exp2}, and \ref{fig:retrieval-curves-exp3}.

\subsection{Head Ablation}
\label{app:head-ablation}

To test for the existence of retrieval heads, which we suspected in the heads with the flatter slopes, we perform additional head ablation experiments. We remove up to three of the most global heads. Results are shown in Table~\ref{table:head-ablations}.

\subsection{Depth Analysis}

Because the Steep strategy underperforms only partially, and the Safe strategy on the other hand is not as strong as expected, we test for the indirect relay of token information across model depth. Results are shown in Table~\ref{table:depth-analysis}.

\begin{table*}
\small
\centering
\begin{tblr}{
  colspec   = {l@{\hspace{8pt}}rr@{\hspace{8pt}}rr@{\hspace{8pt}}rr@{\hspace{8pt}}rrr},
  rowsep    = 2pt,
  colsep    = 2.2pt,
  row{1,2}  = {font=\bfseries},
  column{1} = {leftsep=0pt},
  column{10} = {rightsep=0pt},
}
\toprule
 & \SetCell[c=6]{c} Associative Token Retrieval &  &  &  &  &  & \SetCell[c=3]{c} Downstream &  &  \\\cmidrule[lr]{2-7}\cmidrule[lr]{8-10}
Disabled head(s) & \SetCell[c=1]{c} PK$^{in}$ & \SetCell[c=1]{c} PK$^{out}$ & \SetCell[c=1]{c} NIAH$_s^{in}$ & \SetCell[c=1]{c} NIAH$_s^{out}$ & \SetCell[c=1]{c} NIAH$_m^{in}$ & \SetCell[c=1]{c} NIAH$_m^{out}$ & \SetCell[c=1]{c} CS & \SetCell[c=1]{c} QA & \SetCell[c=1]{c} Language \\
\midrule
None$^*$ & 0.80$_{\pm 0.24}$ & 0.05$_{\pm 0.04}$ & 0.65$_{\pm 0.05}$ & 0.13$_{\pm 0.04}$ & 0.36$_{\pm 0.05}$ & 0.11$_{\pm 0.04}$ & 49.1$_{\pm 1.5}$ & 46.9$_{\pm 0.9}$ & 53.9$_{\pm 0.5}$\\
\midrule
Second last & 0.02$_{\pm 0.02}$ & 0.00$_{\pm 0.00}$ & 0.00$_{\pm 0.00}$ & 0.00$_{\pm 0.00}$ & 0.00$_{\pm 0.00}$ & 0.00$_{\pm 0.00}$ & 48.2$_{\pm 0.7}$ & 45.2$_{\pm 1.8}$ & 49.7$_{\pm 1.6}$\\
Last & 0.34$_{\pm 0.07}$ & 0.00$_{\pm 0.00}$ & 0.07$_{\pm 0.04}$ & 0.00$_{\pm 0.00}$ & 0.04$_{\pm 0.02}$ & 0.00$_{\pm 0.00}$ & 49.2$_{\pm 1.6}$ & 46.3$_{\pm 0.7}$ & 51.3$_{\pm 0.7}$\\
Last two & 0.00$_{\pm 0.00}$ & 0.00$_{\pm 0.00}$ & 0.00$_{\pm 0.00}$ & 0.00$_{\pm 0.00}$ & 0.00$_{\pm 0.00}$ & 0.00$_{\pm 0.00}$ & 48.6$_{\pm 1.7}$ & 43.5$_{\pm 2.1}$ & 47.1$_{\pm 0.5}$\\
Last three & 0.00$_{\pm 0.00}$ & 0.00$_{\pm 0.00}$ & 0.00$_{\pm 0.00}$ & 0.00$_{\pm 0.00}$ & 0.00$_{\pm 0.00}$ & 0.00$_{\pm 0.00}$ & 48.2$_{\pm 2.8}$ & 41.7$_{\pm 1.8}$ & 43.4$_{\pm 0.6}$\\
\midrule
First & 0.00$_{\pm 0.00}$ & 0.00$_{\pm 0.00}$ & 0.00$_{\pm 0.00}$ & 0.00$_{\pm 0.00}$ & 0.00$_{\pm 0.00}$ & 0.00$_{\pm 0.00}$ & 48.5$_{\pm 1.8}$ & 40.2$_{\pm 1.7}$ & 40.8$_{\pm 1.3}$\\
Middle & 0.75$_{\pm 0.27}$ & 0.06$_{\pm 0.07}$ & 0.60$_{\pm 0.05}$ & 0.12$_{\pm 0.05}$ & 0.31$_{\pm 0.08}$ & 0.08$_{\pm 0.05}$ & 48.4$_{\pm 1.3}$ & 45.0$_{\pm 1.5}$ & 51.1$_{\pm 0.5}$\\
\bottomrule
\end{tblr}
\caption{Head ablation testing for retrieval heads. Starting from the ALiBi baseline (*), we disable up to three of the flattest heads (whose slopes range from $2^{-6}$ to $2^{-8}$) across all layers. As a control, we then test with the head in the first and fifth (middle) of the nine heads disabled. Results are averaged over five runs.}
\label{table:head-ablations}
\end{table*}

\begin{table*}
\small
\centering
\begin{tblr}{
  colspec   = {@{}l@{\hspace{8pt}}rr@{\hspace{12pt}}rrrr@{\hspace{12pt}}rrr@{}},
  rowsep    = 2pt,
  colsep    = 3.1pt,
  row{1,2}  = {font=\bfseries},
  column{1} = {leftsep=0pt},
  column{10} = {rightsep=0pt},
}
\toprule
 & \SetCell[c=6]{c} Associative Token Retrieval &  &  &  &  &  & \SetCell[c=3]{c} Downstream &  &  \\\cmidrule[lr]{2-7}\cmidrule[lr]{8-10}
Name & \SetCell[c=1]{c} PK$^{in}$ & \SetCell[c=1]{c} PK$^{out}$ & \SetCell[c=1]{c} NIAH$_s^{in}$ & \SetCell[c=1]{c} NIAH$_s^{out}$ & \SetCell[c=1]{c} NIAH$_m^{in}$ & \SetCell[c=1]{c} NIAH$_m^{out}$ & \SetCell[c=1]{c} CS & \SetCell[c=1]{c} QA & \SetCell[c=1]{c} Language \\
\midrule
Flatter (15L) & 0.58$_{\pm 0.09}$ & 0.00$_{\pm 0.00}$ & 0.57$_{\pm 0.05}$ & 0.06$_{\pm 0.03}$ & 0.32$_{\pm 0.07}$ & 0.05$_{\pm 0.03}$ & 48.0$_{\pm 0.4}$ & \textbf{\underline{47.3$_{\pm 0.5}$}} & 53.1$_{\pm 0.5}$\\
Flat (20L) & 0.61$_{\pm 0.17}$ & 0.00$_{\pm 0.00}$ & 0.59$_{\pm 0.05}$ & 0.09$_{\pm 0.06}$ & 0.30$_{\pm 0.08}$ & 0.06$_{\pm 0.03}$ & 48.5$_{\pm 0.5}$ & 46.5$_{\pm 0.7}$ & 53.4$_{\pm 0.5}$\\
ALiBi (30L)$^*$ & 0.80$_{\pm 0.24}$ & 0.05$_{\pm 0.04}$ & \textbf{0.65$_{\pm 0.05}$} & 0.13$_{\pm 0.04}$ & \textbf{0.36$_{\pm 0.05}$} & \textbf{0.11$_{\pm 0.04}$} & \textbf{49.1$_{\pm 1.5}$} & 46.9$_{\pm 0.9}$ & 53.9$_{\pm 0.5}$\\
Deeper (45L) & \textbf{\underline{0.91$_{\pm 0.10}$}} & \textbf{\underline{0.24$_{\pm 0.34}$}} & 0.61$_{\pm 0.07}$ & \textbf{\underline{0.13$_{\pm 0.09}$}} & 0.30$_{\pm 0.09}$ & 0.09$_{\pm 0.05}$ & 48.5$_{\pm 0.8}$ & \underline{46.9$_{\pm 0.3}$} & \textbf{\underline{54.3$_{\pm 0.8}$}}\\
\bottomrule
\end{tblr}
\caption{Sensitivity analysis over model depth. We take the ALiBi baseline (*) and redistribute the model parameters into flatter and deeper architectures, while approximately keeping the parameter count constant. Results are averaged over five runs.}
\label{table:depth-analysis}
\end{table*}

\begin{table*}
\small
\centering
\begin{tblr}{
  colspec   = {l@{\hspace{9pt}}p{1.5cm}p{1.5cm}p{1.5cm}p{1.5cm}p{1.5cm}p{1.5cm}},
  rowsep    = 2pt,
  colsep    = 11pt,
  row{1,2}  = {font=\bfseries},
  column{1} = {leftsep=0pt},
  column{10} = {rightsep=0pt},
}
\toprule
 & \SetCell[c=6]{c} Commonsense &  &  &  &  &  \\\cmidrule[l]{2-7}
Name & HellaSwag & PIQA & WinoGrande & COPA & SiQA & WSC \\
\midrule
Steep & 32.2$_{\pm 0.4}$ & 62.8$_{\pm 0.5}$ & 50.7$_{\pm 1.5}$ & 61.0$_{\pm 1.7}$ & 36.9$_{\pm 1.0}$ & 36.5$_{\pm 0.0}$\\
Safe & 34.4$_{\pm 0.2}$ & 63.7$_{\pm 0.8}$ & 49.8$_{\pm 1.4}$ & 63.3$_{\pm 2.1}$ & \textbf{\underline{37.7$_{\pm 0.8}$}} & 36.9$_{\pm 0.6}$\\
Wide & 34.8$_{\pm 0.2}$ & 63.8$_{\pm 0.6}$ & 51.1$_{\pm 1.1}$ & 62.0$_{\pm 1.7}$ & \underline{37.4$_{\pm 0.1}$} & 36.5$_{\pm 0.0}$\\
ALiBi & \textbf{35.4$_{\pm 0.2}$} & \textbf{64.2$_{\pm 0.5}$} & \textbf{51.4$_{\pm 0.6}$} & 64.3$_{\pm 0.6}$ & 37.3$_{\pm 0.6}$ & \textbf{44.2$_{\pm 12.5}$}\\
RoPE & 35.3$_{\pm 0.1}$ & 63.4$_{\pm 0.6}$ & 51.3$_{\pm 0.7}$ & \textbf{\underline{64.7$_{\pm 2.1}$}} & \underline{37.5$_{\pm 0.8}$} & 37.8$_{\pm 1.1}$\\
\midrule
Explicit (E) & 35.2$_{\pm 0.3}$ & 63.2$_{\pm 0.5}$ & \textbf{52.4$_{\pm 1.0}$} & 65.0$_{\pm 1.0}$ & \textbf{37.7$_{\pm 0.7}$} & 38.1$_{\pm 2.8}$\\
Clamped (C) & \textbf{35.5$_{\pm 0.3}$} & \textbf{64.1$_{\pm 0.9}$} & 50.8$_{\pm 0.2}$ & 63.7$_{\pm 0.6}$ & 37.7$_{\pm 0.4}$ & \textbf{44.9$_{\pm 14.4}$}\\
Log (L) & 34.1$_{\pm 0.2}$ & 63.3$_{\pm 0.7}$ & 50.9$_{\pm 0.8}$ & 60.7$_{\pm 2.5}$ & 37.1$_{\pm 0.4}$ & 40.7$_{\pm 3.9}$\\
Softcap (S) & 35.2$_{\pm 0.2}$ & 64.0$_{\pm 0.2}$ & 51.9$_{\pm 1.8}$ & \textbf{67.3$_{\pm 3.8}$} & 37.1$_{\pm 0.2}$ & 35.6$_{\pm 1.7}$\\
\midrule
C+E & 35.1$_{\pm 0.1}$ & 63.7$_{\pm 0.2}$ & \textbf{51.9$_{\pm 1.7}$} & 64.0$_{\pm 3.5}$ & \textbf{37.5$_{\pm 0.7}$} & 36.5$_{\pm 0.0}$\\
C+L & 34.1$_{\pm 0.2}$ & 63.3$_{\pm 0.7}$ & 50.9$_{\pm 0.8}$ & 60.7$_{\pm 2.5}$ & 37.1$_{\pm 0.4}$ & 40.7$_{\pm 3.9}$\\
C+S & 35.2$_{\pm 0.1}$ & 64.0$_{\pm 0.8}$ & 50.7$_{\pm 1.6}$ & 64.3$_{\pm 0.6}$ & 37.3$_{\pm 1.0}$ & 37.8$_{\pm 2.2}$\\
E+L & 35.2$_{\pm 0.3}$ & 64.1$_{\pm 0.5}$ & 50.7$_{\pm 0.2}$ & 64.5$_{\pm 2.1}$ & 37.0$_{\pm 0.3}$ & 38.5$_{\pm 2.7}$\\
E+S & \textbf{35.3$_{\pm 0.4}$} & \textbf{64.3$_{\pm 0.7}$} & 50.4$_{\pm 0.5}$ & \textbf{65.0$_{\pm 2.0}$} & 37.5$_{\pm 0.1}$ & 36.5$_{\pm 0.0}$\\
L+S & 34.0$_{\pm 0.1}$ & 62.9$_{\pm 0.5}$ & 50.9$_{\pm 0.4}$ & 63.7$_{\pm 1.5}$ & 37.4$_{\pm 0.1}$ & \textbf{45.2$_{\pm 14.2}$}\\
\midrule
C+E+L & 35.0$_{\pm 0.0}$ & 63.7$_{\pm 0.1}$ & 50.8$_{\pm 0.5}$ & 64.0$_{\pm 1.0}$ & 37.3$_{\pm 0.8}$ & 37.8$_{\pm 2.2}$\\
C+E+S & \textbf{35.4$_{\pm 0.2}$} & \textbf{63.9$_{\pm 1.2}$} & 50.1$_{\pm 1.6}$ & \textbf{65.0$_{\pm 0.0}$} & \textbf{37.7$_{\pm 0.3}$} & 39.1$_{\pm 4.4}$\\
C+L+S & 34.0$_{\pm 0.1}$ & 62.9$_{\pm 0.5}$ & \textbf{51.3$_{\pm 0.4}$} & 62.7$_{\pm 3.2}$ & 37.3$_{\pm 0.3}$ & \textbf{45.2$_{\pm 14.2}$}\\
E+L+S & 35.2$_{\pm 0.2}$ & 63.7$_{\pm 0.5}$ & 50.9$_{\pm 1.7}$ & 64.7$_{\pm 0.6}$ & 37.3$_{\pm 0.9}$ & 39.1$_{\pm 5.3}$\\
\midrule
C+E+L+S & \textbf{35.2$_{\pm 0.2}$} & \textbf{63.6$_{\pm 0.7}$} & \textbf{50.6$_{\pm 1.8}$} & \textbf{65.0$_{\pm 1.0}$} & \textbf{37.5$_{\pm 1.0}$} & \textbf{48.4$_{\pm 14.4}$}\\
\bottomrule
\end{tblr}

\vspace*{0.5cm}

\caption{
\label{table:decoder-benchmarks-cs}
Single results for the decoder benchmark group commonsense (CS).
}
\end{table*}

\begin{table*}
\small
\centering
\SetTblrInner{rowsep=1pt,colsep=1pt}
\begin{tblr}{
  colspec   = {l@{\hspace{6pt}}p{1.5cm}p{1.5cm}p{1.5cm}p{1.5cm}p{1.5cm}@{\hspace{6pt}}p{1.7cm}p{1.5cm}p{1.3cm}},
  rowsep    = 2pt,
  colsep    = 2pt,
  row{1,2}  = {font=\bfseries},
  column{1} = {leftsep=0pt},
  column{8} = {leftsep=2pt},
  column{10} = {rightsep=0pt},
}
\toprule
 & \SetCell[c=5]{c} Question Answering &  &  &  &  & \SetCell[c=3]{c} Language &  &  \\\cmidrule[lr]{2-6}\cmidrule[lr]{7-9}
Name & ArcE & ArcC & BoolQ & SciQ & Race & LAMBADA & Blimp & WiC \\
\midrule
Steep & 48.2$_{\pm 1.5}$ & 26.2$_{\pm 0.2}$ & 59.0$_{\pm 2.5}$ & 67.5$_{\pm 1.0}$ & 29.1$_{\pm 0.2}$ & 24.9$_{\pm 0.4}$ & \textbf{82.1$_{\pm 0.5}$} & 50.0$_{\pm 1.5}$\\
Safe & 49.6$_{\pm 0.2}$ & \textbf{27.0$_{\pm 0.9}$} & 57.7$_{\pm 5.6}$ & 71.0$_{\pm 1.6}$ & 29.9$_{\pm 0.4}$ & 28.1$_{\pm 0.7}$ & 81.1$_{\pm 0.8}$ & 50.1$_{\pm 0.2}$\\
Wide & \textbf{50.7$_{\pm 0.6}$} & 25.9$_{\pm 0.3}$ & \textbf{61.5$_{\pm 0.5}$} & 71.3$_{\pm 0.6}$ & 30.5$_{\pm 0.2}$ & 29.1$_{\pm 0.5}$ & 81.9$_{\pm 0.0}$ & 50.0$_{\pm 0.0}$\\
ALiBi & 50.4$_{\pm 0.9}$ & 26.3$_{\pm 1.1}$ & 54.6$_{\pm 7.1}$ & \textbf{71.4$_{\pm 1.1}$} & 30.5$_{\pm 0.4}$ & 29.9$_{\pm 0.9}$ & 81.7$_{\pm 0.9}$ & 50.5$_{\pm 0.4}$\\
RoPE & 49.5$_{\pm 0.8}$ & 26.9$_{\pm 1.5}$ & 58.5$_{\pm 3.5}$ & 68.9$_{\pm 0.9}$ & \textbf{30.7$_{\pm 0.1}$} & \textbf{29.9$_{\pm 0.7}$} & 81.6$_{\pm 0.5}$ & \textbf{52.0$_{\pm 1.8}$}\\
\midrule
Explicit (E) & 49.6$_{\pm 0.6}$ & 26.4$_{\pm 0.5}$ & \textbf{60.3$_{\pm 1.6}$} & \textbf{70.7$_{\pm 0.7}$} & 30.1$_{\pm 0.5}$ & 29.3$_{\pm 1.2}$ & 81.7$_{\pm 0.1}$ & 50.1$_{\pm 0.1}$\\
Clamped (C) & 49.7$_{\pm 0.5}$ & 27.2$_{\pm 0.4}$ & 59.0$_{\pm 2.9}$ & 70.3$_{\pm 1.8}$ & 30.5$_{\pm 1.0}$ & \textbf{29.7$_{\pm 0.8}$} & 81.7$_{\pm 0.4}$ & 50.3$_{\pm 0.2}$\\
Log (L) & 48.3$_{\pm 1.2}$ & 26.8$_{\pm 0.4}$ & 58.4$_{\pm 2.3}$ & 69.5$_{\pm 0.1}$ & \textbf{31.1$_{\pm 0.1}$} & 27.3$_{\pm 1.0}$ & 81.2$_{\pm 0.3}$ & \textbf{50.5$_{\pm 0.5}$}\\
Softcap (S) & \textbf{49.9$_{\pm 0.6}$} & \textbf{27.2$_{\pm 0.2}$} & 57.6$_{\pm 4.7}$ & 70.5$_{\pm 0.8}$ & 30.6$_{\pm 0.3}$ & 29.4$_{\pm 0.4}$ & \textbf{81.7$_{\pm 0.8}$} & 50.1$_{\pm 0.2}$\\
\midrule
C+E & 49.8$_{\pm 0.5}$ & 26.7$_{\pm 0.8}$ & 60.9$_{\pm 0.9}$ & 70.0$_{\pm 0.9}$ & 30.3$_{\pm 0.9}$ & \textbf{29.6$_{\pm 0.8}$} & 81.8$_{\pm 0.3}$ & 50.0$_{\pm 0.2}$\\
C+L & 48.3$_{\pm 1.2}$ & 26.8$_{\pm 0.4}$ & 58.4$_{\pm 2.3}$ & 69.5$_{\pm 0.1}$ & 31.1$_{\pm 0.1}$ & 27.3$_{\pm 1.0}$ & 81.2$_{\pm 0.3}$ & 50.5$_{\pm 0.5}$\\
C+S & 49.3$_{\pm 1.5}$ & 26.2$_{\pm 0.6}$ & 60.0$_{\pm 2.1}$ & 70.2$_{\pm 1.1}$ & 30.1$_{\pm 1.0}$ & 29.1$_{\pm 0.1}$ & 81.9$_{\pm 0.3}$ & 49.9$_{\pm 0.1}$\\
E+L & 49.2$_{\pm 0.1}$ & 26.1$_{\pm 1.0}$ & 60.5$_{\pm 0.4}$ & 70.4$_{\pm 1.2}$ & 29.9$_{\pm 0.3}$ & 28.8$_{\pm 0.5}$ & 82.1$_{\pm 0.2}$ & 50.5$_{\pm 0.7}$\\
E+S & \textbf{50.6$_{\pm 1.2}$} & \textbf{26.9$_{\pm 0.7}$} & \textbf{61.0$_{\pm 0.6}$} & \textbf{70.9$_{\pm 0.7}$} & \textbf{31.1$_{\pm 0.7}$} & 29.4$_{\pm 1.1}$ & \textbf{82.3$_{\pm 0.4}$} & 50.0$_{\pm 0.0}$\\
L+S & 47.7$_{\pm 0.7}$ & 26.0$_{\pm 0.7}$ & 58.0$_{\pm 2.9}$ & 68.5$_{\pm 0.4}$ & 30.7$_{\pm 1.1}$ & 28.3$_{\pm 0.4}$ & 81.0$_{\pm 0.2}$ & \textbf{52.8$_{\pm 3.6}$}\\
\midrule
C+E+L & \textbf{49.9$_{\pm 1.0}$} & 26.5$_{\pm 1.0}$ & \textbf{61.0$_{\pm 0.3}$} & \textbf{70.8$_{\pm 1.0}$} & 30.4$_{\pm 0.6}$ & 28.6$_{\pm 0.2}$ & 81.9$_{\pm 0.4}$ & 50.4$_{\pm 0.6}$\\
C+E+S & 49.3$_{\pm 0.9}$ & 27.3$_{\pm 0.6}$ & 59.1$_{\pm 2.4}$ & 70.6$_{\pm 0.6}$ & 30.2$_{\pm 0.8}$ & \textbf{28.9$_{\pm 0.7}$} & \textbf{82.2$_{\pm 0.3}$} & 50.2$_{\pm 0.2}$\\
C+L+S & 47.6$_{\pm 0.8}$ & 25.7$_{\pm 0.6}$ & 57.7$_{\pm 2.6}$ & 68.5$_{\pm 0.5}$ & \textbf{30.5$_{\pm 1.5}$} & 28.1$_{\pm 0.1}$ & 81.0$_{\pm 0.2}$ & \textbf{52.9$_{\pm 3.5}$}\\
E+L+S & 49.3$_{\pm 1.2}$ & \textbf{27.4$_{\pm 0.3}$} & 60.8$_{\pm 1.2}$ & 70.5$_{\pm 1.5}$ & 30.3$_{\pm 1.0}$ & 28.6$_{\pm 0.4}$ & 81.4$_{\pm 0.7}$ & 49.9$_{\pm 0.1}$\\
\midrule
C+E+L+S & \textbf{49.7$_{\pm 1.2}$} & \textbf{27.5$_{\pm 0.1}$} & \textbf{60.4$_{\pm 1.3}$} & \textbf{70.3$_{\pm 1.7}$} & \textbf{29.9$_{\pm 1.4}$} & \textbf{28.7$_{\pm 0.4}$} & \textbf{81.3$_{\pm 0.6}$} & \textbf{50.2$_{\pm 0.3}$}\\
\bottomrule
\end{tblr}

\caption{
\label{table:decoder-benchmarks-qa-lg}
Single results for the decoder benchmark groups question answering (QA) and language (LG).
}
\end{table*}

\end{document}